%% file: main.tex
\documentclass[final,3p,times]{elsarticle}
\usepackage[english]{babel}
\usepackage{bm}                                 
\usepackage{amsmath,amsfonts,amssymb} 
\usepackage{graphicx}                 
\usepackage{siunitx}                  
\usepackage[table,xcdraw,dvipsnames]{xcolor}  
\usepackage{textcomp}                 
\usepackage{lmodern,babel,adjustbox,booktabs,multirow,wrapfig,listings,pdflscape,enumitem,longtable,makecell,verbatim} 
\usepackage[font=normalsize,labelfont=sf,textfont=sf,labelsep=period]{subfig,caption} 
\usepackage{hyperref}
\usepackage[capitalize]{cleveref} 

\usepackage{algorithm,algorithmic}   
\usepackage[ruled,vlined,algo2e]{algorithm2e}
\usepackage{stfloats}                
\usepackage{url}                     
\usepackage{pifont}                  
\usepackage{indentfirst}             
\usepackage{epstopdf}                
\usepackage{balance}                 
\usepackage{microtype}               
\usepackage[normalem]{ulem}   

\newcommand{\model}{MorphoSeg\xspace}

\input{preamble}

\journal{Neurocomputing Journal}

\begin{document}

\begin{frontmatter}



\title{\model: An Uncertainty-Aware Deep Learning Method for Biomedical Segmentation of Complex Cellular Morphologies}

\author[1]{Tianhao Zhang}
\author[2]{Heather J. McCourty}
\author[2]{Berardo M. Sanchez-Tafolla}
\author[2]{Anton Nikolaev}
\author[1]{Lyudmila S. Mihaylova}

\affiliation[1]{organization={School of Electrical and Electronic Enigeering, The University of Sheffield}, 
            city={Sheffield},
            country={United Kingdom}}

\affiliation[2]{organization={School of Biosciences, The University of Sheffield}, 
            city={Sheffield},
            country={United Kingdom}}


\begin{abstract}
Deep learning has revolutionized medical and biological imaging, particularly in segmentation tasks. However, segmenting biological cells remains challenging due to the high variability and complexity of cell shapes. Addressing this challenge requires high-quality datasets that accurately represent the diverse morphologies found in biological cells. Existing cell segmentation datasets are often limited by their focus on regular and uniform shapes. In this paper, we introduce a novel benchmark dataset of Ntera-2 (NT2) cells, a pluripotent carcinoma cell line, exhibiting diverse morphologies across multiple stages of differentiation, capturing the intricate and heterogeneous cellular structures that complicate segmentation tasks. To address these challenges, we propose an uncertainty-aware deep learning framework for complex cellular morphology segmentation (\model) by incorporating sampling of virtual outliers from low-likelihood regions during training. Our comprehensive experimental evaluations against state-of-the-art baselines demonstrate that \model significantly enhances segmentation accuracy, achieving up to a 7.74\% increase in the Dice Similarity Coefficient (DSC) and a 28.36 reduction in the Hausdorff Distance. These findings highlight the effectiveness of our dataset and methodology in advancing cell segmentation capabilities, especially for complex and variable cell morphologies. The dataset and source code is publicly available at \url{https://github.com/RanchoGoose/MorphoSeg}.

\end{abstract}



\begin{keyword}
%
%
Biomedical Segmentation, Cell Segmentation, Machine Learning, Deep Learning, Ntera-2 Cells, Data Repository, Complex Cell Shapes, Vision Transformer
\end{keyword}

\end{frontmatter}


\input{sec/1_intro}

\input{sec/2_background}

\input{sec/3_method}
\input{sec/4_data}
\input{sec/5_exp}

\input{sec/6_conclusion}

\bibliographystyle{elsarticle-num}
\bibliography{main}

\end{document}

%% file: preamble.tex
\renewcommand{\paragraph}[1]{\vspace{0.2em}\noindent \textbf{#1 \hspace{0.2em}}}

\def \d1{\mathds{1}}

\def \d1{\mathds{1}}

\def\1{\mathbf{1}}


%% file: sec/1_intro.tex
\section{Introduction}
The advent of deep learning has significantly advanced the field of medical imaging, particularly in segmentation tasks. While substantial progress has been made in segmenting well-defined objects such as organs in computed tomography (CT) scans~\cite{chen2024transunet, ronneberger2015u_unet0, zhang2023customized, zhao20213d}, cell segmentation presents unique challenges due to the complexity and variability of cell shapes~\cite{saha2018her2net, stringer2021cellpose, 616aacc35244ab9dcb31d91f, livecell}. Existing datasets for cell segmentation~\cite{israel2023foundation, livecell, 10.1371/journal.pcbi.1005177, Moen803205} often fall short in covering the full spectrum of cell types, hindering the effective training of deep learning models for diverse research applications.

In the broader context of medical image analysis, cell segmentation is crucial yet challenging, particularly due to the high variability in cell stages and the resolutions of medical images. Current state-of-the-art (SOTA) deep learning models, including advanced architectures such as Convolutional Neural Networks (CNNs)~\cite{stringer2021cellpose, li2017h_unet1} and Transformers~\cite{chen2024transunet, zhang2023customized}, typically require large, well-annotated datasets, which are resource-intensive to acquire. Beyond focusing on model algorithms, some researchers have shifted towards developing methods for rapid, easy data annotation or advancing weak, semi, or unsupervised learning algorithms to address data insufficiency~\cite{5fe31675d4150a363c80bcfa, 63634e0790e50fcafd9fca3a}. However, irregular cell shapes and the need for multi-stage observation
remain significant challenges.

\begin{figure}
    \centering
    \includegraphics[width=0.58\linewidth]{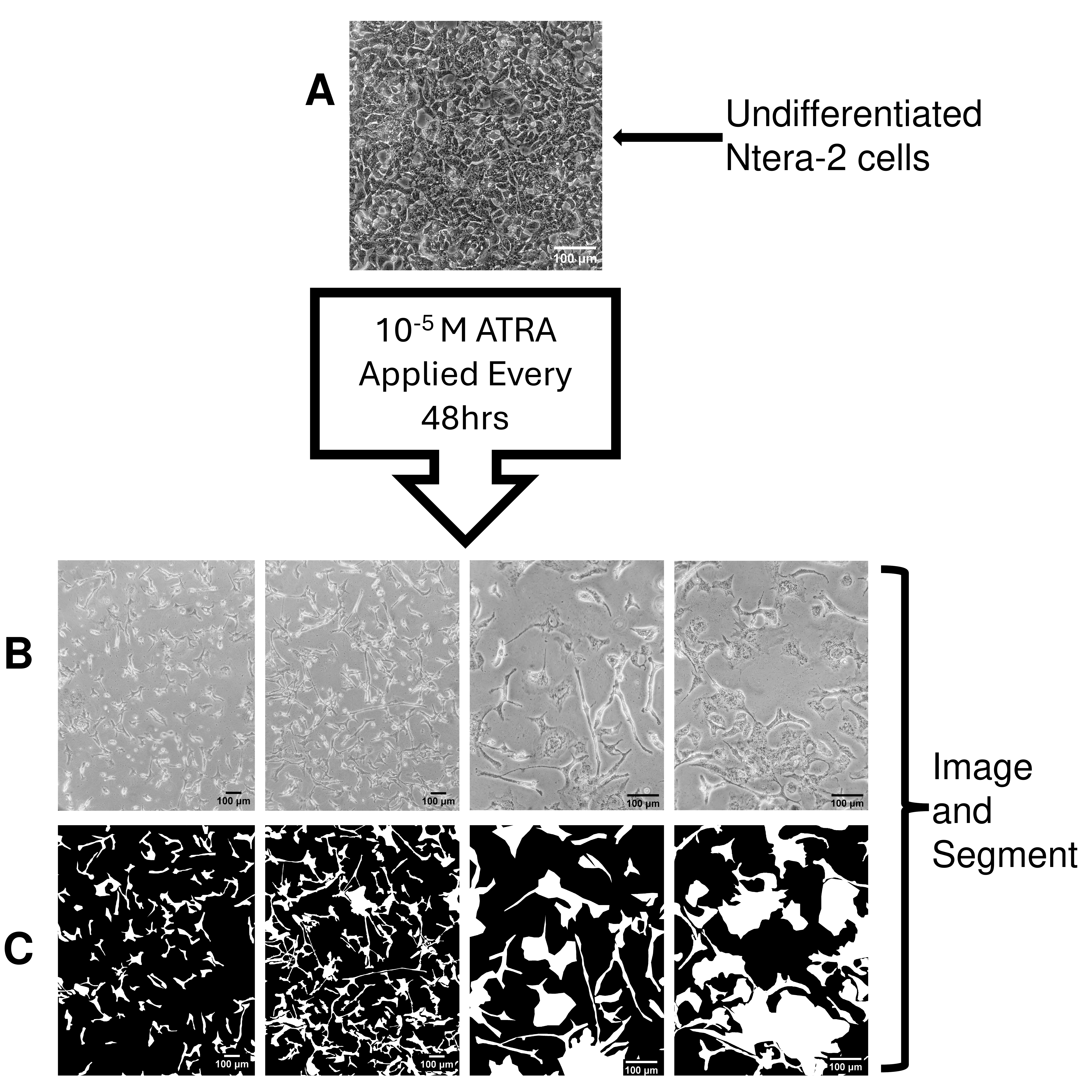}
    \caption{Undifferentiated and Differentiating NT2 Cell Example Images From the Training Set. 
    (A) Undifferentiated Ntera-2 cells. 
    (B) shows the original bright-field images of cells exposed to $10^{-5}$ M ATRA every 48 hours for 2 weeks. (C) displays the mask created from manual annotations. The images highlight the diverse data set used for training.}
    \label{fig:cell}
\end{figure}

In this paper, we present a novel benchmark dataset featuring  Ntera-2 cells (NT2) cells~\cite{data_repository:2024:sheffield}. NT2 is a pluripotent carcinoma cell line known for its ability to differentiate into neuro-ectodermal lineages~\cite{pleasure1993}. NT2 cells have previously been shown to transform into neurons, after exposure to all-\textit{trans}-retinoic acid~\cite{andrews1984} (ATRA).  During differentiation, NT2 cells exhibit a range of heterogeneous morphologies, as shown in Figure~\ref{fig:cell}, which poses challenges for reliable segmentation. The dataset includes images of NT2 cells at numerous stages of differentiation, spanning from day 7 to day 12, and captures the diverse and complex cell morphologies formed during this period. The complexities can make it difficult to accurately annotate cells, with some cellular features barely discernible even to trained experts. Furthermore, the variability in shape across different stages complicates the segmentation process, necessitating robust deep-learning solutions capable of adapting to such heterogeneity.


To address the aforementioned challenges, we propose a novel uncertainty-aware enhancement algorithm based on the SOTA medical segmentation framework TransUNet~\cite{chen2024transunet}. Our approach involves sampling virtual outliers from the low-likelihood regions of the class-conditional distributions during the training stage to improve the decision boundaries of the segmented objects. This strategy enhances the robustness of the deep learning models, making them effective solutions in such complex scenarios. By incorporating uncertainties into the training objective, our proposed deep learning model achieves strong segmentation of cells with random shapes. When compared against several benchmarks, we demonstrate significant improvements with our proposed dataset: the mean Dice Similarity Coefficient (DSC) increased from 80.35\% to 86.57\% and the mean Hausdorff Distance (HD95) decreased from 21.98 to 15.75. These results confirm that our approach not only mitigates data scarcity but also enhances the model's generalization capabilities across varied cell morphologies. We have provided the full code and pre-trained models with simple inference instructions to facilitate easy implementation and further research by field researchers.

The key contributions of this paper are:
\begin{enumerate}
    \item \textbf{Development of a Novel Cell Repository:} We introduce a pluripotent carcinoma cell repository specifically curated to support the development and validation of advanced biological cell segmentation models. Unlike conventional datasets that predominantly feature cells with regular, round morphologies, our dataset encompasses cells with complex morphologies and transitional stages, presenting a more challenging segmentation task.
    
    \item \textbf{An Innovative Cell Segmentation Algorithm MorphoSeg:} We propose MorphoSeg, a novel approach designed to address the challenges posed by our dataset. Our method leverages synthetic outliers from low-likelihood regions in the latent space to significantly enhance the training performance of state-of-the-art cell segmentation models, especially in scenarios with limited data availability. This approach demonstrates superior performance compared to existing methods that often struggle under similar conditions.
    
    \item \textbf{Open-Source Implementation and Comprehensive Evaluation:} We release the complete code and detailed data processing procedures as open-source, enabling researchers in the biological field to readily apply our method to raw images and obtain segmentation masks directly. Our extensive quantitative and qualitative evaluations confirm that the proposed model outperforms current techniques.
\end{enumerate}

The remainder of this paper is organized as follows: Section~\ref{sec:back} reviews the literature on deep learning methods for biological cell segmentation, and outlier generation methods. Section~\ref{sec:method} details our proposed model's methodology and framework. Section~\ref{sec:data} elaborates on the dataset and preprocessing techniques. Section~\ref{sec:exp} presents experimental setup, results, analyses, and ablation studies, as well as inference and visualization techniques. Finally, Section~\ref{sec:con} concludes the paper, summarizing key findings and future research directions.

%% file: sec/2_background.tex
\section{Related Work} \label{sec:back}
\subsection{Medical and Biological Image Segmentation.} 
In recent years, deep learning techniques in computer vision have achieved promising performance in medical image segmentation. Encoder-decoder-based CNN structures have been widely applied, with U-Net and its variants significantly advancing the field~\cite{ronneberger2015u_unet0, li2017h_unet1, liu20173d_unet2, milletari2016v_unet3, yu2017volumetric_unet4}. Building upon this structure, UNet++ incorporates densely connected links to enhance the fusion of multi-scale features, improving the analysis of medical images across CT, MRI, and EM modalities~\cite{zhou2019unet++_unet7}. Furthermore, nnU-Net, a self-configuring U-Net, has shown remarkable performance across various segmentation tasks with minimal manual parameter tuning~\cite{isensee2021nnu}. In terms of segmenting cell structures, a framework named Cellpose representing the CNN-based Unet serires performance with the enhancement . The Cellpose model utilizes a two-headed output architecture to predict cell probabilities and spatial organization, effectively handling cells of regular shapes~\cite{stringer2021cellpose}.

Semantic segmentation is pivotal for clinical tasks such as disease detection, differential diagnosis, survival prediction, therapy planning, and treatment response assessment. Besides deep learning methods based on CNN structures, attention mechanisms~\cite{vaswani2017attention} are extensively employed in semantic segmentation to prioritize salient features. The integration of Vision Transformers (ViTs) into medical segmentation networks has enhanced performance by leveraging self-attention mechanisms~\cite{dosovitskiy2020image}. Some research focuses on combining the strengths of transformers with traditional CNN architectures, creating hybrid encoder structures. TransUNet combines CNNs for spatial information extraction with transformers for capturing long-range dependencies, featuring a U-Net structure with self-attention mechanisms to ensure comprehensive image analysis~\cite{chen2024transunet}. This integration makes TransUNet particularly effective for detailed medical and biological image segmentation, where preserving spatial hierarchies is crucial. Building upon this concept, SwinUNet~\cite{swinunet} incorporates Swin Transformer's shifted-window approach, effectively capturing fine-grained features in small-scale targets. Additionally, ScaleFormer~\cite{Huang2022ScaleFormerRT} further enhances performance by adopting a more powerful Transformer-based mixer, demonstrating superior capability in handling complex segmentation tasks.

Other research includes employing channel attention to capture boundary-aware features for enhancing polyp segmentation~\cite{fan2020pranet}. CellViT replaces the CNN encoder with a transformer block in the U-Net architecture~\cite{CellViT}. The recent SAMed model extends the capabilities of the Segment Anything Model (SAM)~\cite{kirillov2023segment} to medical imaging by employing a Low-rank Adaptation (LoRA) fine-tuning strategy to adapt the large-scale image encoder for medical contexts effectively~\cite{zhang2023customized}. This method illustrates the advantage of using extensive pre-trained models for specialized medical segmentation tasks. However, SAM is designed for multi-class segmentation, posing limitations for binary segmentation scenarios.

Regarding on complex cell structure cell segmentation, some recent works have tried to use geometric-aware approaches. Topological consistency is critical in segmentation tasks, especially for reticular images where even slight changes can significantly affect downstream analysis \cite{gupta2024topology}. For curvilinear structures such as vasculature and road networks, a method leveraging discrete Morse theory has been proposed to capture structural uncertainties and identify error-prone connections \cite{lin2024topout}. A topology-aware loss function that combines skeleton-based weighting with boundary rectification further enhances segmentation accuracy by emphasizing topologically critical regions \cite{liu2024enhancing}. In digital pathology, a semi-supervised approach employs a noise-aware topological consistency loss to learn robust representations from unlabeled images, reducing errors in gland and nuclei segmentation \cite{xu2025semi}. In contrast to these methods, our approach integrates uncertainty estimation directly into the training loss via an outlier synthesis technique, making it compatible with and complementary to existing topological segmentation frameworks.



\subsection{Virtual Outliers Synthesis}  
Outlier detection plays a crucial role in data mining and statistics, where outliers are defined as abnormalities or deviations from expected patterns \cite{Aggarwal2017}. In deep learning, these outliers are often interpreted as anomalies relative to inlier data, and various methods have been developed to identify them \cite{wang2019progress}. Unlike traditional approaches that focus on outlier detection, recent works in computer vision have explored synthesizing outliers to enhance the accuracy and robustness of deep learning models, particularly in scenarios where datasets are insufficient, such as the medical image classification~\cite{boukerche2020outlier}.

Synthetic data has also been employed to quantify model uncertainty, especially for out-of-distribution (OOD) detection. One study leverages synthetic data to better define the boundaries between known and unknown samples \cite{du2022towards}, which is particularly relevant for detecting unseen cells or cells at transitional stages. OOD detection methods often struggle with the subtle differences among foreground elements in medical images \cite{dong2022neural}. 

Synthesis methods have shown promise in various domains, yet their application to medical and biological image segmentation remains relatively underexplored, with few approaches targeting deep learning model training. A recent work combines image synthesis, Gaussian mixture models, and one-class support vector machines to robustly detect hyperintense lesions in cerebral small vessel disease without being limited by pathology or location \cite{BOWLES2017643}. Another research introduces a synthetic data generation pipeline, SinGAN-Seg, which leverages a single training image to produce realistic medical images and corresponding masks, thereby enhancing segmentation performance in scenarios with limited data \cite{thambawita2022singan}. Additionally, an adversarial confidence learning framework integrates voxel- and region-wise confidence information into supervised segmentation and synthesis models, resulting in improved visual perception and quantitative precision on clinical datasets \cite{Nie2020}. 

In contrast to these methods, our proposed approach dynamically estimates class-conditional Gaussian distributions during training, shaping the uncertainty surface over time with a specially formulated loss. In our cell segmentation task, segmentation targets—specifically NT2 cells at various developmental stages—are treated as OOD data. The demonstrated effectiveness of virtual outlier synthesis in enhancing OOD detection \cite{du2022towards} motivates our approach to incorporate synthesized outliers during training to improve segmentation accuracy.

%% file: sec/3_method.tex
\begin{figure}
    \centering
    \includegraphics[width=1\linewidth]{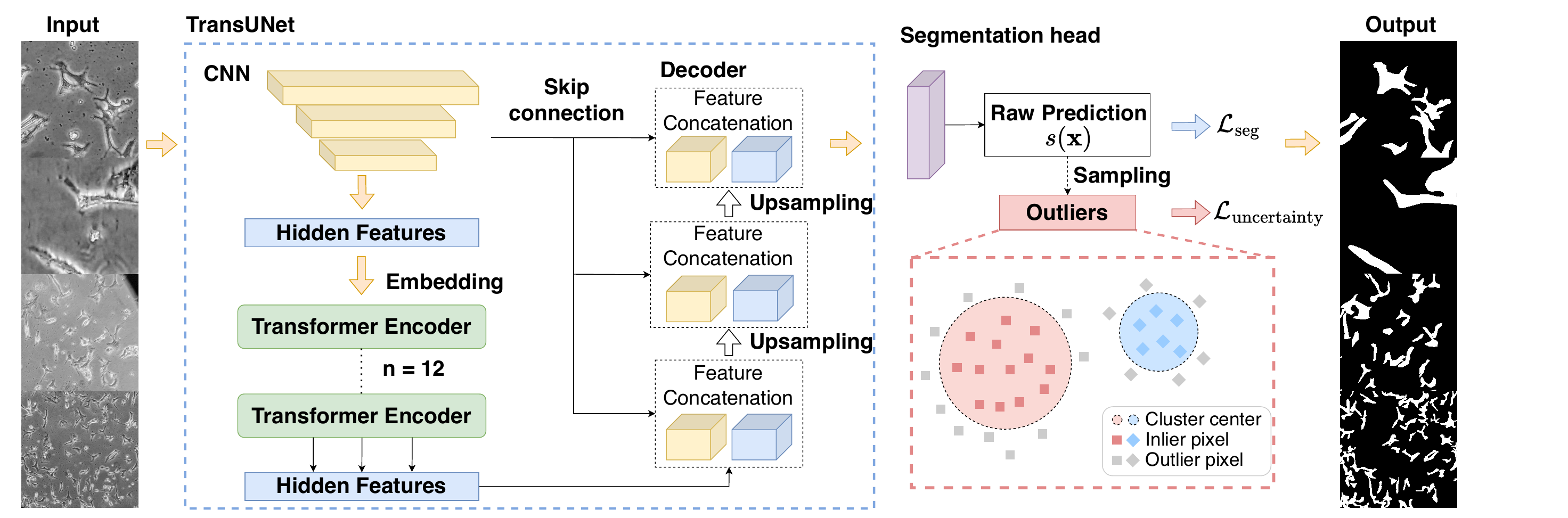}
    \caption{Overview of the \model. MorphoSeg employs the TransUNet architecture~\cite{chen2024transunet} as an example. The red dotted lines illustrate the virtual outlier sampling process, where feature representations are modeled as class-conditional Gaussians and virtual outliers $v$ are sampled from regions of low likelihood. These outliers contribute to an uncertainty loss for regularization, and the synthetic segmentation map—with an uncertainty estimation branch $\mathcal{L}_\text{uncertainty}$—is trained jointly with the mixed segmentation loss $\mathcal{L}_\text{seg}$ as defined in Eq.~(\ref{eq:seg_loss}).}
    \label{fig:overall}
\end{figure}

\section{Methodology} \label{sec:method}
In this section, we detail our proposed approach and outline the overall model architecture. First, Section~\ref{sec:problem} provides background information and an overview of the TransUNet framework for image segmentation. Next, Section~\ref{sec:gda} describes our enhancement of cell segmentation via virtual outlier sampling. Finally, Section~\ref{sec:training} introduces the uncertainty-aware training objective used in MorphoSeg.

\subsection{TransUNet for Cell Segmentation} \label{sec:problem}
Given an image $\bm{\mathrm{x}} \in \mathbb{R}^{H \times W \times C}$ with a spatial resolution of $H \times W$ and $C$ channels, our objective is to predict the corresponding pixel-wise label map of the same dimensions. For a pure Transformer encoder structure, following~\cite{dosovitskiy2020image}, we first perform tokenization by reshaping the input $\bm{\mathrm{x}}$ into a sequence of flattened 2D patches \{$\bm{\mathrm{x}}^i_p \in \mathbb{R}^{P^2 \cdot C}|i=1,..,N\}$, where each patch is of size $P \times P$ and $N=\frac{HW}{P^2}$ is the number of image patches (\emph{i.e.}, the input sequence length). We map the vectorized patches $\bm{\mathrm{x}}_p$ into a latent $D$-dimensional embedding space using a trainable linear projection.
To encode the patch spatial information, we learn specific position embeddings which are added to the patch embeddings to retain positional information as follows:



\begin{align}
    \bm{\mathrm{z}}_0 &= [\bm{\mathrm{x}}^1_p \bm{\mathrm{E}}; \, \bm{\mathrm{x}}^2_p \bm{\mathrm{E}}; \cdots; \, \bm{\mathrm{x}}^{N}_p \bm{\mathrm{E}} ] + \bm{\mathrm{E}}_{pos}, \label{eq:embedding} 
\end{align}

\noindent where $\bm{\mathrm{E}} \in \mathbb{R}^{(P^2 \cdot C) \times D}$ is the patch embedding projection, and $\bm{\mathrm{E}}_{pos}  \in \mathbb{R}^{N \times D}$ denotes the position embedding. 

The concatenated sequence $\mathbf{T}$ is processed through the Transformer encoder layers to yield the final representations. Then, we employ multi-headed self-attention (MSA)~\cite{vaswani2017attention} (Eq.~\ref{eq:MSA}) and multi-layer perceptron (MLP) blocks (Eq.~\ref{eq:MLP}). Layer normalization (LN) is applied before each block (Eq.~\ref{eq:final_representation}), as described in the following equations:
\begin{align}
    \mathbf{Z}_0 &= \mathbf{T}, \\
    \mathbf{Z}^\prime_{\ell} &= \text{MSA}(\text{LN}(\mathbf{Z}_{\ell-1})) + \mathbf{Z}_{\ell-1}, \quad \ell = 1 \ldots L, \label{eq:MSA} \\ 
    \mathbf{Z}_{\ell} &= \text{MLP}(\text{LN}(\mathbf{Z}^\prime_{\ell})) + \mathbf{Z}^\prime_{\ell}, \quad \ell = 1 \ldots L,  \label{eq:MLP} \\
    \mathbf{Y} &= \text{LN}(\mathbf{Z}_L[0]),  \label{eq:final_representation}
\end{align}

\noindent where $L$ denotes the number of Transformer layers, and $\mathbf{Z}_L[0]$ represents the final layer's class token representation. The output $\mathbf{Y} \in \mathbb{R}^{D}$ serves as the input to a classifier head for the task at hand. For segmentation, an effective approach is to upsample the encoded feature representation $\bm{\mathrm{z}}_L \in \mathbb{R}^{\frac{HW}{P^2} \times D}$ to the full resolution. This involves reshaping the encoded feature from $\frac{HW}{P^2}$ to $\frac{H}{P} \times \frac{W}{P}$, applying a $1 \times 1$ convolution to reduce the channel size to the number of classes, and then bilinearly upsampling the feature map to the full resolution $H \times W$ for the final segmentation outcome. 

Due to the limited dimensions of ViT input tokens, input image patch sizes $\frac{H}{P} \times \frac{W}{P}$ are often much smaller than the original image resolution $H \times W$ in high-resolution medical images, leading to a loss of low-level details (e.g., cell shape and boundary). TransUNet addresses this by using a CNN-Transformer hybrid model, where a CNN first extracts a feature map from the input. Patch embedding is then applied to $1 \times 1$ patches from the CNN feature map instead of raw images, ensuring better retention of details. We present the hybrid encoder in Figure~\ref{fig:overall} because it yields better performance, as demonstrated in the comparative analysis of different encoders presented in Table~\ref{tab:main_results} in Section~\ref{sec:exp_results}.

\vspace{1ex}\noindent\textbf{U-Net Structure and Skip Connections.}  
Following TransUNet, we adopt the standard U-Net design with skip connections, replacing naive upsampling with a Cascaded Upsampler (CUP). After reshaping the feature sequence $\bm{\mathrm{z}}_L \in \mathbb{R}^{\frac{HW}{P^2} \times D}$ into a $\frac{H}{P} \times \frac{W}{P} \times D$ feature map, CUP employs multiple upsampling blocks to progressively restore full resolution. Each block comprises a $2\times$ upsampling operator, a $3 \times 3$ convolution, and a ReLU activation, enabling effective multi-scale feature aggregation.


Figure~\ref{fig:overall} illustrates the overall framework of our approach.  One thing to notice is that our method is not highly relied on the TransUNet as the outliers sampling processes lies in the latent space and invarient to the model structure, we present the TransUNet here simply as it represents the Transformer based Unet Structure model framework for cell segmentations. The details of how raw high-resolution images are subdivided into small patches to satisfy the input constraints of the vision transformer and to capture fine cellular details presented in Section~\ref{sec:data_pre}.

\subsection{Robust Cell Segmentation via Virtual Outlier Sampling} \label{sec:gda}

Our proposed NT2 cell dataset presents two major challenges. First, high-resolution medical images often contain small, densely packed cells that are highly sensitive to pixel-level variations, making accurate human annotation particularly difficult. Second, there is a lack of sufficient training data for the diverse stages of pluripotent carcinoma cells, which limits the model’s ability to generalize across different developmental stages.

To address these issues, we propose a \emph{virtual outlier sampling} strategy that is applied exclusively during training. Instead of directly sampling outliers from the input image space $\bm{\mathrm{x}} \in \mathbb{R}^{H \times W \times C}$, we synthesize virtual outliers in the latent feature space $\bm{z} \in \mathbb{R}^{H' \times W' \times C'}$, where $H'$, $W'$, and $C'$ differ from $H$, $W$, and $C$ due to the spatial and channel transformations introduced by the encoder network.

Operating in the latent space provides several benefits. First, it reduces computational complexity due to the lower dimensionality of latent representations. Second, it enhances training stability and model robustness, as learning to handle synthetic perturbations in the latent space encourages the model to better capture the underlying data distribution. In contrast, perturbing the input space directly may lead to unstable training and degraded performance.

This strategy also enables the incorporation of an \emph{uncertainty loss} term, $\mathcal{L}_{\text{uncertainty}}$, which guides the model to account for potential outlier regions during training, further improving generalization and segmentation accuracy under limited and imbalanced data conditions.

Specifically, we opt to sample virtual outliers directly from the output logits of the segmentation map, denoted by $s(\mathbf{x}_i)$, rather than from earlier latent representations. This strategy not only reduces computational load but also minimizes the risk of training instabilities, such as loss explosion, that could arise from discrepancies between artificially added noise and true label distributions.

Given the generic nature of our method, we establish general notations applicable to similar tasks in image classification or segmentation. We model the features of all target classes as conditional multivariate Gaussian distributions:
\begin{align} 
p_\theta(s(\mathbf{x}) | y=k) = \mathcal{N}(\boldsymbol{\mu}_k, \boldsymbol{\Sigma}), \label{eq:distribution}
\end{align}

\noindent where $\boldsymbol{\mu}_k$ represents the mean for class $k \in \{1, 2, \dots, K\}$, and $\boldsymbol{\Sigma}$ is the covariance matrix shared across classes. The function $s(\mathbf{x}) \in \mathbb{R}^m$ represents the segmentation map of a cell instance $\mathbf{x}$, where $m$ is significantly smaller than the input dimension $d$. In our specific application to the pluripotent carcinoma cell dataset, we focus on two classes: the cells and the background, implying $k=2$.

To estimate the parameters of the Gaussian distribution, we calculate the empirical class means and covariance matrix from the training samples $\{\mathbf{x}_i, y_i\}_{i=1}^{N}$:
\begin{align}
    \widehat{\boldsymbol{\mu}}_k &= \frac{1}{N_k} \sum_{i: y_i=k} s(\mathbf{x}_i), \\
    \widehat{\boldsymbol{\Sigma}} &= \frac{1}{N} \sum_{k=1}^K \sum_{i: y_i=k} \left(s(\mathbf{x}_i) - \widehat{\boldsymbol{\mu}}_k\right) \left(s(\mathbf{x}_i) - \widehat{\boldsymbol{\mu}}_k\right)^\top.
\end{align}

where $N_k$ is the number of objects in class $k$, and $N$ is the total number of objects. We use online estimation for efficient training, where we maintain a class-conditional queue with $|Q_k|$ object instances from each class. In each iteration, we enqueue the embeddings of objects to their corresponding class-conditional queues and dequeue the same number of object embeddings.



\noindent\textbf{Sampling from the Feature Representation Space} We propose sampling virtual outliers from the feature representation space of cell images, utilizing the multivariate distributions estimated from these features. These virtual outliers are designed to refine the decision boundary between different stages of cell development and other outlier data, which may not be adequately represented in the training set.

To accomplish this, we generate virtual outliers $\mathcal{V}_k$ from the $\epsilon$-likelihood region of the estimated class-conditional distribution for each cell stage:
\begin{align}
  \mathcal{V}_k = \left\{ v_k \mid \frac{1}{(2 \pi)^{m / 2}|\widehat{\Sigma}|^{1 / 2}} \exp \left(-\frac{1}{2}(v_k-\widehat{\mu}_k)^\top \widehat{\Sigma}^{-1}(v_k-\widehat{\mu}_k)\right) < \epsilon \right\},
   \label{eq:virtual}
\end{align}
where $v_k \sim \mathcal{N}(\widehat{\mu}_k, \widehat{\Sigma}_k)$ represents the sampled virtual outliers for class $k$. These samples are chosen from a sublevel set defined by a small $\epsilon$, ensuring that the outliers are near the decision boundary of the class. With the virtual outliers $v_k$ sampled from the predicted segmentation map $s(\mathbf{x}_i)$, we effectively capture the inherent uncertainty of the segmentation process. In particular, the set of virtual outliers, $\mathcal{V}_k$, is integrated into the segmentation pipeline to update $s(\mathbf{x})$, yielding an uncertainty-enhanced output that we denote as $f_{\text{out}}(\mathbf{x}; \theta)$. This updated output is then utilized in our uncertainty-aware loss functions in the $\mathcal{L}_{\text{Dice\_out}}$ and $\mathcal{L}_{\text{CE\_out}}$ in Eq.~\ref{eq:ce_vos_loss} and ~\ref{eq:dice_vos_loss}.

\begin{figure}
    \centering
    \subfloat[Original cell images from the training set]{
        \includegraphics[width=0.95\textwidth]{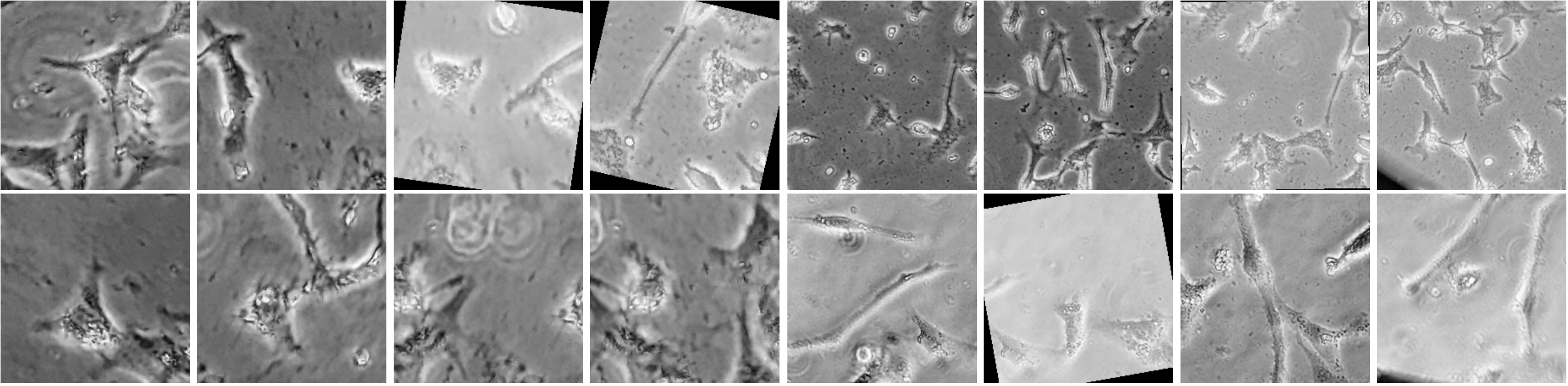}
    } 
    \hfill
    \subfloat[Corresponding ground truth masks]{
        \includegraphics[width=0.95\textwidth]{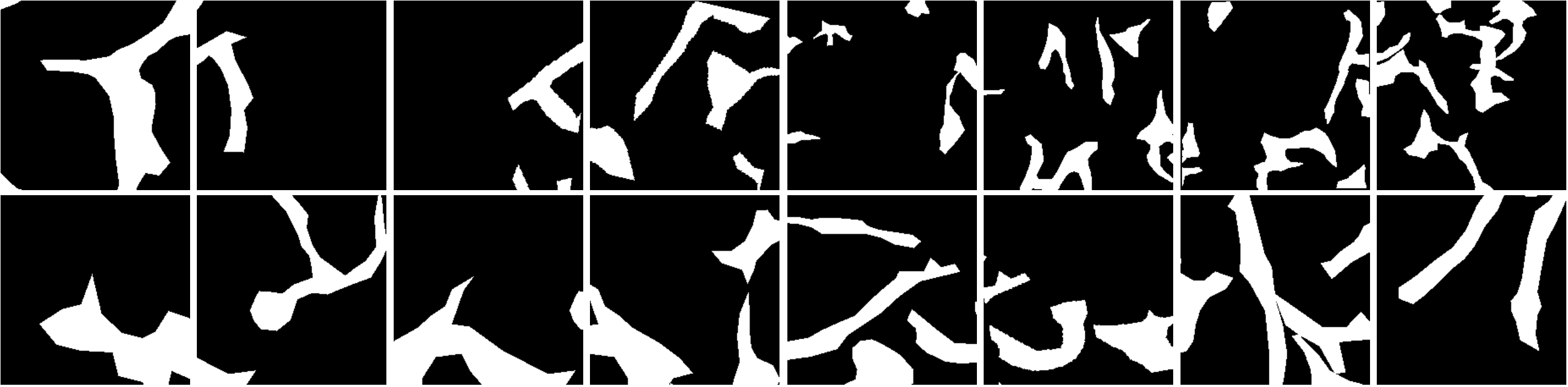}
    } 
    \hfill
    \subfloat[Visualization of the proposed segmentation map with synthesis outliers]{
        \includegraphics[width=0.95\textwidth]{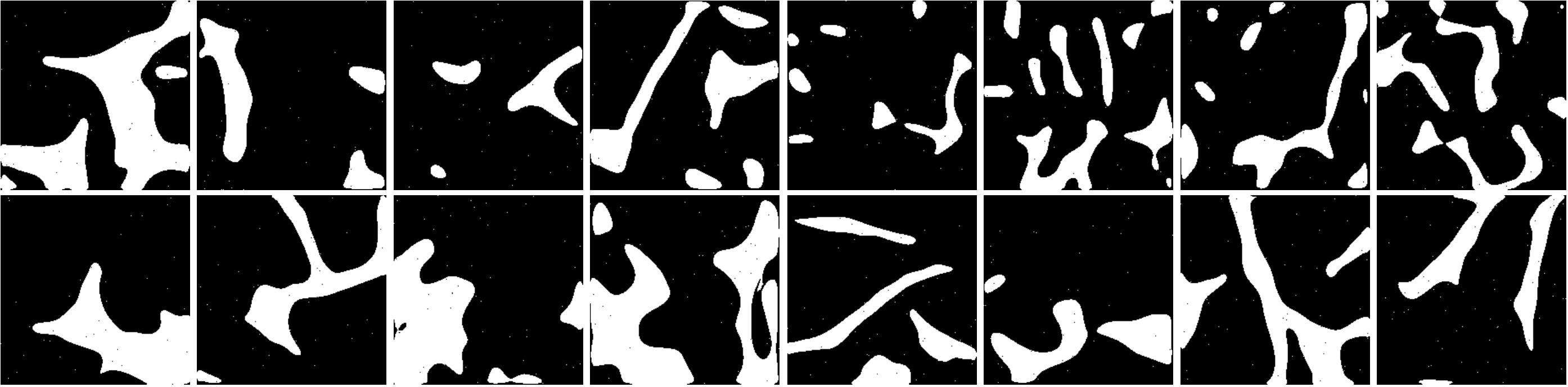}
    } 
    \caption{Examples showcasing the training with visual outliers. Patches are taken from the training set, presented in sizes of \(224 \times 224\) and \(448 \times 448\). Compared to the ground truth masks, the synthetic outlier samples simulate small cell dots at the pixel level, which are challenging for human annotation. Additionally, small white dots are presented as outliers, which may be difficult to discern without magnification.}
    \label{fig:vos_vis}
\end{figure}

We hypothesize that most of the cell pixels are near the boundary of existing cells as shown in Figure~\ref{fig:vos_vis}. Compared to the ground truth masks shown in the second subfigure in Figure~\ref{fig:vos_vis}, the synthesized outliers contain more information about the cells, which are difficult for humans to annotate due to their small size. This strategy aims to enhance the model's sensitivity to subtle yet critical variations in cell morphology that might indicate different developmental stages or pathological conditions.

\subsection{Uncertainty-aware Training Objective}
\label{sec:training}
The objective is to develop an uncertainty-aware deep learning model capable of predicting a segmentation map \( \hat{S} \) for input cell images. To this end, we introduce a new overall training objective, which utilizes virtual outliers as discussed in Section~\ref{sec:gda}. We start by defining the standard Dice loss between the predicted pixel scores \( \hat{y} \) and the ground truth targets \( y \):
\begin{align}
\mathcal{L}_{\text{Dice}}(f(\mathbf{x}; \theta), y) = 1 - \frac{2 \sum_{i} \hat{y}_i y_i + \epsilon}{\sum_{i} \hat{y}_i^2 + \sum_{i} y_i^2 + \epsilon},
\label{eq:dice_loss}
\end{align}
where \( \hat{y}_i \) represents the predicted probabilities for the actual class labels \( y_i \), and \( \epsilon \) is a small constant introduced to prevent division by zero.

In the case of medical segmentation~\cite{5a260c8117c44a4ba8a30b08}, Dice loss is frequently combined with Cross Entropy loss to form the overall training objective. The Cross Entropy loss is defined as follows:
\begin{align}
\mathcal{L}_{\text{CE}}(f(\mathbf{x}; \theta), y) &= -\sum_{i} y_i \log(f(\mathbf{x}_i; \theta)),
\label{eq:ce_loss}
\end{align}
where \(f(\mathbf{x}_i; \theta)\) is the model's output for the input \(\mathbf{x}_i\), and \(y_i\) is the corresponding true label. 

The overall training objective for medical segmentation, as used in vanilla TransUNet, combines both Cross Entropy and Dice losses~\cite{tian2024fairseg}. This combined loss function is employed as the primary training loss, defined as:
\begin{align}
\mathcal{L}_\text{seg} = \lambda_1 \mathcal{L}_{\text{CE}}(\hat{S}_l, D(S)) + \lambda_2 \mathcal{L}_{\text{DICE}}(\hat{S}_l, D(S)).
\label{eq:seg_loss}
\end{align}
The weights \(\lambda_1\) and \(\lambda_2\) are used to balance these two terms, typically set to be equally distributed~\cite{chen2024transunet}.

With the virtual outliers $v_k$ sampled from the predicted segmentation map $s(\mathbf{x}_i)$, we effectively capture the inherent uncertainty of the segmentation process. To make our model uncertainty-aware, we adapt both the Dice loss and Cross Entropy loss for training as follows:

\begin{align}
\mathcal{L}_{\text{Dice\_out}}(f_{\text{out}}(\mathbf{x}; \theta), y) &= 1 - \frac{2 \sum_{i} \hat{y}_{\text{syn}, i} y_i + \epsilon}{\sum_{i} \hat{y}_{\text{syn}, i}^2 + \sum_{i} y_i^2 + \epsilon}, \label{eq:dice_vos_loss} \\
\mathcal{L}_{\text{CE\_out}}(f_{\text{out}}(\mathbf{x}; \theta), y) &= -\sum_{i} y_i \log(f_{\text{out}}(\mathbf{x}_i; \theta)),
\label{eq:ce_vos_loss}
\end{align}
where \( \hat{y}_{\text{syn}, i} \) represents the predicted probabilities for the actual class labels \( y_i \) from the synthetic outputs, and \( f_{\text{out}}(\mathbf{x}_i; \theta) \) denotes the model's output for synthetic data designed to simulate under-represented features.

Furthermore, we model the uncertainty captured by the virtual outliers through a specific loss function:
\begin{align}
\mathcal{L}_\text{uncertainty} = \beta_1 \mathcal{L}_{\text{CE\_out}}(\hat{S}_l, D(S)) + \beta_2 \mathcal{L}_{\text{DICE\_out}}(\hat{S}_l, D(S)),
\label{eq:uncer_loss}
\end{align}
where \(\beta_1\) and \(\beta_2\) are coefficients that weigh the contributions of the cross-entropy and Dice losses, respectively, in quantifying the uncertainty associated with the segmentation of cells. 

\noindent\textbf{Overall Training Objective.} Although the proposed loss function is initially designed for binary segmentation, it is adaptable for multi-class segmentation scenarios. Our overall training objective integrates the segmentation loss, which includes both Dice and Cross Entropy losses, and a regularization component that addresses model uncertainty:
\begin{align}
    \min_{\theta} \mathbb{E}_{(\mathbf{x}, y) \sim \mathcal{D}} \left[ \lambda \mathcal{L}_{\text{seg}}(\mathbf{x}, y; \theta) + \beta \mathcal{L}_{\text{uncertainty}}(\mathbf{x}, y; \theta) \right],
    \label{eq:all_loss}
\end{align}
where $\lambda$ and \(\beta\) is the weighting coefficient for the uncertainty regularization. The segmentation loss, \(\mathcal{L}_{\text{seg}}\), combines the Dice loss, \(\mathcal{L}_{\text{DICE}}\), and the cross-entropy loss, \(\mathcal{L}_{\text{CE}}\), formulated to effectively handle the segmentation of cells across different stages and morphologies. The uncertainty component, \(\mathcal{L}_{\text{uncertainty}}\), leverages synthetic data to enhance the model’s ability to predict underrepresented features and quantify predictive confidence. Ablation studies, detailed in Section~\ref{sec:abl_loss}, demonstrate the efficacy of this loss function in improving segmentation accuracy and model robustness.

%% file: sec/4_data.tex
\section{Dataset} \label{sec:data}

\subsection{Data Description} \label{sec:data_des} NT2 cells were exposed to $10\textsuperscript{-5}$ ATRA every 48 hours for 14 days ~\cite{pleasure1993, andrews1984}. The cells were imaged using a bright-field microscope on treatment days. ATRA exposure induces neuronal differentiation in NT2 cells, leading to notable morphological changes including an increase in cytoplasm size, elongation of the cell body and neurite extension, as illustrated in Figure~\ref{fig:cell}B. 

The NT2 dataset consists of 105 high-resolution raw images of NT2 cells undergoing differentiation, of which 36 images are fully annotated, including 2425 individual cells annotated. These images have resolutions primarily around $4000 \times 3000$ pixels, enabling detailed morphological analysis and accurate segmentation.


\subsection{Data Preprocessing} \label{sec:data_pre}
\noindent\textbf{Data Augmentation.} 
We randomly partition the entire dataset into two subsets, allocating 80\% for training and 20\% for testing. To effectively train our TransUNet-based MorphoSeg model, we augment the data by partitioning each high-resolution image into smaller, manageable patches. Specifically, we extract patches of sizes $224 \times 224$, $448 \times 448$, $1000 \times 1000$, $1500 \times 1500$, and $2000 \times 2000$ pixels using an overlapping strategy with a 35\% overlap. This approach preserves the spatial context across adjacent patches, which is essential for capturing the complex spatial relationships inherent in cell structures. In addition to these patches, the original full-resolution images and their corresponding masks are included to form the complete training and testing dataset.

Using this technique, we generate a total of 15,710 patches. This distribution results in 12,568 patches for training and 3,142 patches for testing. To optimize the model's performance, we implement a data pruning strategy during both the training and testing stages, excluding all patches that do not contain any part of the annotated masks, which will reduce the training set to 10,016 samples for training.

In addition to evaluating performance on the patch-based testing set, we also assess our model on full-resolution images that lack annotations (i.e., the 69 images not included in the 36 annotated images out of 105 total). This evaluation more closely reflects real-world scenarios where researchers directly apply the trained model to raw images, yielding segmentation masks for the entire image. Visualization examples of these results are presented in Figure~\ref{fig:cell_vis}.

%% file: sec/5_exp.tex
\section{Experiments} \label{sec:exp}
\subsection{Implementation details} \label{sec:exp_setup}
We evaluate our model against established benchmarks including Cellpose and TransUNet, utilizing the AdamW optimizer with exponential learning rate decay on the R50-ViT-B architecture. Specifically for Cellpose, we utilize two of the highest performing pre-trained configurations, cyto3 and cyto2, both set with a cell diameter of $30$. These models are trained with a learning rate of $0.01$, a weight decay of $5 \times 10^{-5}$, and over $200$ epochs. The batch size for training is set at $8$, with verbose output enabled for detailed progress tracking. During testing, we adapted the Cellpose model to handle larger cells by setting the diameter parameter to $100$, with no flow threshold and a cell probability threshold of $0.2$, adjusting the channels accordingly to match specific imaging conditions. For the TransUNet, we adhere which includes two types of encoders: a pure Transformer-based encoder, denoted as ``ViT-B16,'' and a hybrid encoder that combines ResNet-50 with ViT, denoted as ``R50-B16''. The input resolution and patch size \(P\) are set to 224$\times$224 and 16, respectively. We set the learning rate to 0.01, momentum to 0.9, and weight decay to \(1 \times 10^{-4}\). All models are trained over 200 epochs, with sampling commencing at the 150th epoch. Batch sizes of 128 and 256 are implemented depending on the model size, focusing on 10,0000 critical samples by selectively sampling 1,0000 pixels per image to enhance training efficiency and effectiveness.

\subsection{Evaluation Metrics}
We evaluate our method using four key metrics: the average Dice Similarity Coefficient (DSC)~\cite{shamir2019continuous} for assessing overlap accuracy, the 95th Percentile Hausdorff Distance (HD95)~\cite{huttenlocher1993comparing} for measuring boundary alignment, the Intersection over Union (IoU)~\cite{rezatofighi2019generalized} for quantifying region overlap, and the mean Average Precision (mAP)~\cite{henderson2017end} for overall object detection performance. These metrics, calculated from true positives, false positives, and false negatives, offer a comprehensive view of segmentation quality, while qualitative evaluations further validate the practical effectiveness of our approach.

\begin{table}
    \centering
    \scalebox{0.8}{

\input{tab/main_results}}
    \caption{Comparison on our proposed cell dataset. All methods incorporating outlier synthesis are trained with a sample size of $100,000$, and the selection count is set to $10,000$. $\uparrow$ indicates that larger values are preferable, while $\downarrow$ indicates that smaller values are better. All numerical values are presented as percentages. \textbf{Bold} numbers denote the best performance, with results presented as mean $\pm$ standard deviation over 3 runs.}
    \label{tab:main_results}
\end{table}

\subsection{Comparison with Baselines} \label{sec:exp_results}
Table~\ref{tab:main_results} presents a comprehensive comparison of our proposed approach, MorphoSeg, against several baselines including Cellpose, TransUNet, and TopoSeg UNet++ , SwinUNet  and ScaleFormer. Cellpose, a CNN-based U-Net model tailored for cell segmentation, is evaluated in its cyto3 and cyto2 variants. Although Cellpose benefits from requiring fewer annotations, its performance is hindered by its reliance on cell flow and diameter estimations—characteristics that vary significantly across different cell stages. This variability is reflected in its DSC scores of $63.66$ (cyto3) and $48.15$ (cyto2), indicating limitations when applied to datasets with diverse cell morphologies.

In addition to Cellpose, we integrated MorphoSeg with another CNN-based architecture, UNet++~\cite{zhou2019unet++_unet7}. To maintain consistent experimental settings, we sampled the virtual outliers at the output level of the segmentation map rather than at a reduced dimensionality in the latent space. Although sampling in the latent space might simplify training and further improve results, even under these suboptimal conditions, our MorphoSeg still yielded a measurable improvement for UNet++, as demonstrated in Table~\ref{tab:main_results}.

Furthermore, we extended MorphoSeg to two state-of-the-art Transformer-based segmentation backbone networks, SwinUNet~\cite{swinunet} and ScaleFormer~\cite{Huang2022ScaleFormerRT}. The results reported in Table~\ref{tab:main_results} were obtained using the official default settings for these models. The integration of MorphoSeg led to significant performance gains across multiple evaluation metrics for both SwinUnet and ScaleFormer, thereby demonstrating the robustness and versatility of our approach.

TransUNet was evaluated in two variants: ViT-B16 and R50-B16. The ViT-B16 variant exhibits suboptimal performance with a DSC of only $25.00$, likely due to challenges in distinguishing cells from the background under varying lighting conditions. In contrast, the R50-B16 variant demonstrates markedly better performance, achieving a DSC of $80.35$. This improvement underscores the benefit of incorporating a ResNet-50-based encoder for effectively capturing low-level features in high-resolution medical images.

We implement TopoSeg based on TransUNet. As a state-of-the-art semi-supervised method that leverages topological consistency, it shows moderate performance. Its ViT-B16 variant attains a DSC of $26.53$, while the R50-B16 variant improves to $66.14$. However, across all metrics, TopoSeg remains less competitive compared to the best-performing models. The configurations used for TopoSeg remain the default settings.

Our proposed method, MorphoSeg, further enhances segmentation accuracy by incorporating virtual outlier sampling and an uncertainty-aware training objective. When combined with the R50-B16 encoder, MorphoSeg achieves a DSC of $86.57$, a HD95 of $15.75$, and superior IoU and mAP scores compared to all baselines. These results highlight the effectiveness of our approach in managing the inherent variability in cell morphology. Additionally, while an integration of our method with TopoSeg was explored using both ViT-B16 and R50-B16 variants, the standalone performance of MorphoSeg with the R50-B16 encoder remains the most competitive, emphasizing its robust capability for accurate cell segmentation.

\begin{table}
    \centering
    \scalebox{0.8}{

\input{tab/abl_loss}}
     \caption{Ablation study on the effect of different loss function designs for MorphoSeg. All methods incorporating outlier synthesis are trained using the same configuration as in Table~\ref{tab:main_results}. $\uparrow$ indicates that larger values are preferable, while $\downarrow$ indicates that smaller values are better. \textbf{Bold} numbers denote the best performance, with results presented as mean $\pm$ standard deviation over 3 runs.}
    \label{tab:abl_loss}
\end{table} 

\begin{table}
    \centering
    \small
    \scalebox{0.8}{

\input{tab/abl_layers}}
     \caption{Ablation studies on different network architectures. For each encoder variant, we report the number of Transformer layers, attention heads, and indicate whether the encoder is hybrid (ResNet backbone; $\checkmark$) or pure Transformer ($\times$). Results are presented for three loss variants: Balance, Norm, and Pareto.}
    \label{tab:abl_layers}
\end{table} 

\subsection{Ablation Studies}  \label{sec:exp_abl}
\noindent\textbf{Incorporating Uncertainty into Loss Function Design.} \label{sec:abl_loss}
To determine the optimal values for the weight parameters $\lambda$ and $\beta$ in Equation.(\ref{eq:all_loss}), we evaluate three sophisticated approaches to weight the loss components, tailored for enhanced model training and testing. The \textbf{Original Balancing} method, consistent with TranUNet, evenly distributes weights across all loss components, simplifying optimization but potentially overlooking specific task requirements within the model. Alternatively, the \textbf{Normalized Losses} method dynamically scales each loss component by its magnitude:
\begin{equation}
\mathcal{L}_{\text{norm}} = \sum_{i} \frac{\mathcal{L}_i}{\| \mathcal{L}_i \|} + \epsilon,
\label{eq:scaled_loss}
\end{equation}
where $\mathcal{L}_i$ denotes individual loss components, $\|\cdot\|$ signifies the norm operation applied to each loss component to normalize its scale, and $\epsilon$ is a small constant to prevent division by zero. This adjustment ensures that hyperparameters $\lambda_i$ are iteratively tuned to maintain balance among the loss components. A refinement of this approach normalizes subsequent losses relative to the first, calculated as:
\begin{equation}
\mathcal{L}_{\text{pareto}} = \mathcal{L}_{\text{loss}} + \sum_{i \neq \text{loss}} \frac{\mathcal{L}_i}{\| \mathcal{L}_i / \mathcal{L}_{\text{loss}} \|},
\label{eq:refined_loss}
\end{equation}
which potentially provides a more rational scaling by relating all losses back to the primary loss component. Lastly, the \textbf{Pareto Optimization} strategy~\cite{NeurIPS2018_Sener_Koltun}, aims to achieve a Pareto efficient solution by treating each loss as an objective in a multi-objective optimization framework, wherein improvements in one objective are sought without significant compromises in others, facilitating an equilibrium that optimizes overall model performance.

\begin{table}
    \centering
    \scalebox{0.6}{

\input{tab/dataset}}
    \caption{Segmentation results on the BloodCell and LIVECell datasets of several benchmarks. The results of using MorphoSeg are reported with three loss functions: Balance, Norm, and Pareto. The TransUNet presented in this table uses the hybrid encoder of R50-ViT-B-16. Best results for each dataset for each method is marked in \textbf{Bold} numbers.}
    \label{tab:dataset}
\end{table} 

Table~\ref{tab:abl_loss} shows an ablation study comparing three loss strategies for MorphoSeg using two encoder backbones, ViT-B16 and R50-B16. With ViT-B16, the Balance loss yields slightly higher DSC and lower HD95 while all methods register low IoU values at high thresholds. In contrast, the R50-B16 encoder achieves much better performance overall and the Pareto strategy delivers the best results with the highest DSC and mAP and the lowest HD95. These results suggest that the choice of loss is more influential with R50-B16 and that the Pareto approach offers a modest advantage.

\begin{figure}[hp]
    \centering
    \includegraphics[width=0.95\linewidth]{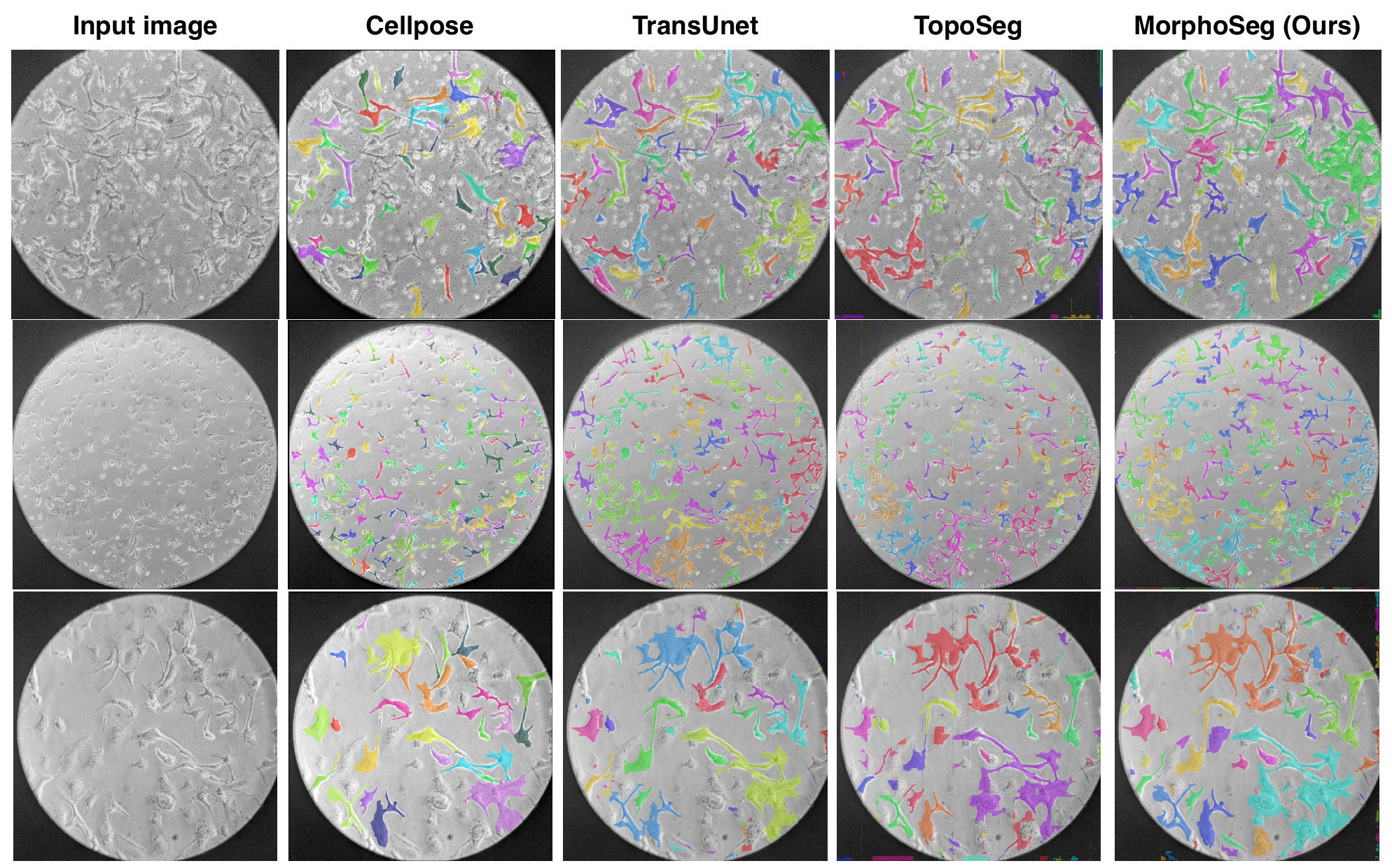}
    \caption{Visualization of full input images with model outputs. The presented images are the direct outputs from the model without annotations or ground truth masks, representing unseen data during training and testing. Due to the high resolution, differences among the outputs of TransUNet, TopoSemi, and \model are not easily discernible. For a more detailed comparison, zoomed-in results are provided in Figure~\ref{fig:vis}.}
    \label{fig:cell_vis}
\end{figure}

\noindent\textbf{Ablation Studies on Different Network Architectures.} \label{sec:abl_layers}
To evaluate the effect of different network components, we performed experiments on various architectures with differing numbers of Transformer layers and attention heads. Table~\ref{tab:abl_layers} summarizes these results, which show that larger networks generally yield better performance. Specifically, the ``R50-B16'' model represents our hybrid architecture combining a ResNet-50 backbone with a Vision Transformer configured in a base setting, consisting of 12 Transformer layers with 12 self-attention heads per layer and a patch size of 16. In contrast, the ``R50-L32'' model uses the same ResNet-50 backbone but incorporates a larger Transformer module with 32 layers and 16 self-attention heads per layer. For the pure Vision Transformer architectures, ViT-B16 comprises 12 layers with 12 attention heads per layer, while ViT-L16 features 24 layers with 16 attention heads per layer.

\subsection{Experiments on Additional Datasets} \label{sec:other_dataset}
To further validate the generalization and robustness of our proposed MorphoSeg, we performed experiments on two widely recognized datasets: the BloodCell dataset \cite{Ljosa2012} and the LIVECell dataset \cite{livecell}. The BloodCell dataset (BBBC041v1) consists of 1,364 microscopic images of Giemsa-stained human blood smears infected with \textit{Plasmodium vivax}, containing approximately 80,000 cells. The LIVECell dataset includes 5,239 expert-annotated phase-contrast microscopy images, comprising 1,686,352 annotated cells across eight cell types, with predefined splits for training, validation, and testing, along with various subset sizes for scalability experiments. 

Noteably that both the BloodCell and LIVECell datasets present segmentation tasks that are relatively simpler compared to our primary dataset due to the consistent cellular morphologies observed. Despite these variations, our MorphoSeg method achieves promiss performance improvements, as summarized in Table~\ref{tab:dataset}, highlighting its effectiveness across diverse cellular imaging conditions.

\subsection{Patch-Based Approach for High-Resolution Inference and Visualization} \label{sec:exp_vis}
Given the inherent limitations of Vision Transformers, where token size constraints can result in vague and imprecise inference on large images, we propose an advanced patch-based inference technique to address the sensitivity to pixel-level cellular details. This approach enhances segmentation accuracy for high-resolution biological cell images while making the framework more practical and accessible for researchers. Example visualizations of the results are presented in Figure~\ref{fig:cell_vis}.

\begin{figure}[hp]
    \centering
    \includegraphics[width=0.55\linewidth]{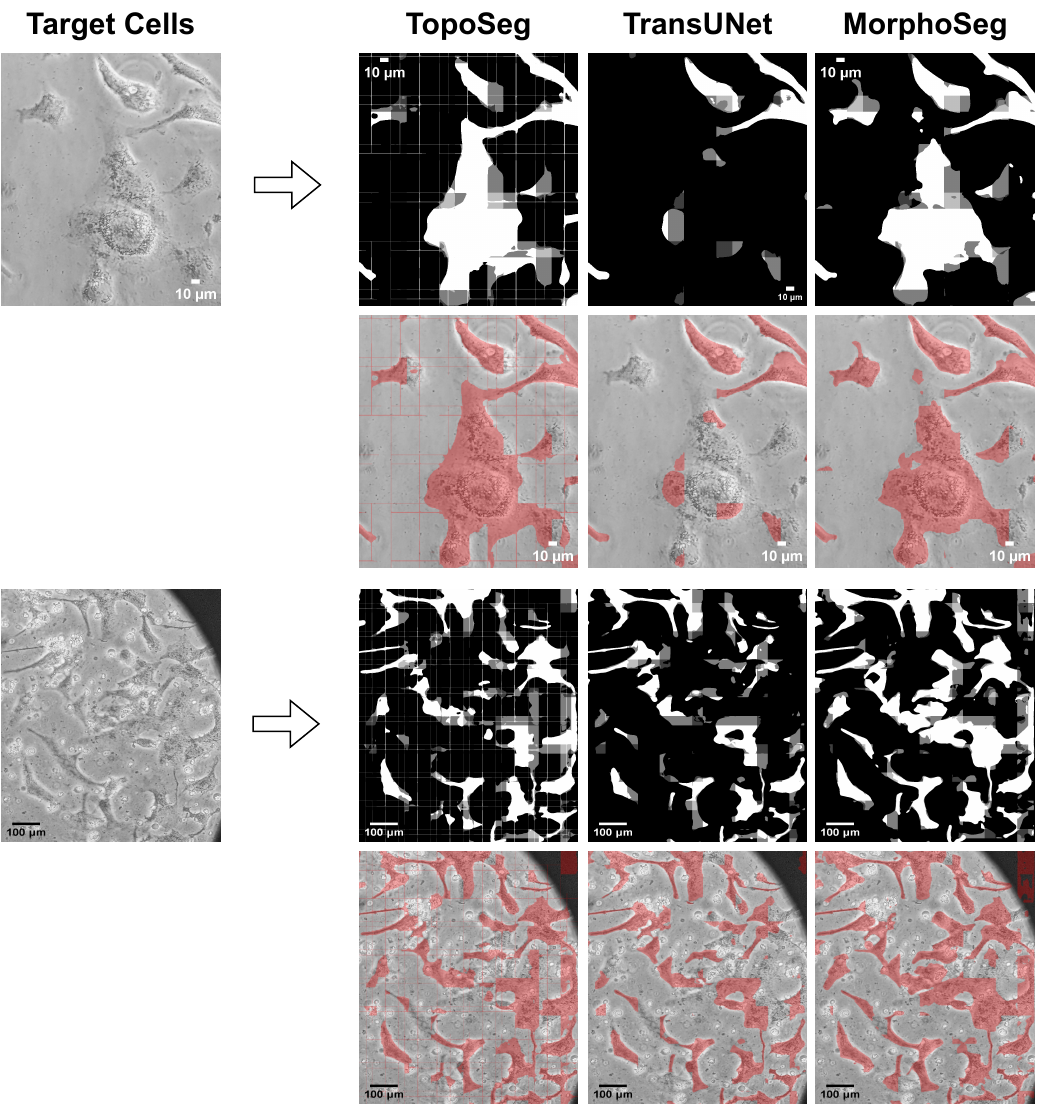}
    \caption{Zoomed-in visualization of Ntera-2 cell segmentation results on the testing dataset, including target cell annotations, predicted segmentation masks, and corresponding overlay outputs. As a detailed supplement to Figure~\ref{fig:cell_vis}, this comparison highlights the performance of \model against TopoSeg and TransUNet. Notable differences between segmentation methods are observed, with some regions exhibiting incomplete segmentation, while others merge multiple cells into a single segmentation.}
    \label{fig:vis}
\end{figure}

During a single image inference stage, the image is initially subdivided into patches of size $224 \times 224$, similar to the data augmentation strategy presented in Section~\ref{sec:data_pre}. This subdivision allows the model to focus on and accurately detect fine-grained variations at the boundaries of the Ntera-2 cells and their surrounding environment. Each patch undergoes a separate inference process. To preserve the global spatial context, we incorporate an overlapping strategy with a $56$-pixel margin. This overlap ensures continuity and minimizes edge artifacts. The inference results from the patches are then combined using an average pooling strategy and visualized as segmentation masks, shown in red in Figure~\ref{fig:cell_vis} and Figure~\ref{fig:vis}. This end-to-end training and visualization strategy significantly facilitates the application of our framework, enabling researchers to use it without requiring further modifications.

Figure~\ref{fig:cell_vis} presents inference results with segmented masks overlaid on the original raw images. For comparison, we also show outputs from Cellpose, TransUNet, and TopoSeg using the same input images. It is evident that Cellpose fails to detect a significant number of cells, while TopoSeg produces unsatisfactory results with many spurious, thin lines. In contrast, MorphoSeg yields results comparable to TransUNet but with noticeable improvements in the fine details. Although the high resolution of the raw images can mask these subtle differences in the full-view images, the zoomed-in regions in Figure~\ref{fig:vis} clearly demonstrate the enhanced segmentation quality. These qualitative findings are consistent with the quantitative results reported in Table~\ref{tab:main_results}.

%% file: tab/main_results.tex
\begin{tabular}{l c c c c c c c c}
\toprule
\multicolumn{1}{l}{\textbf{Method}} & \multicolumn{2}{c}{\textbf{Model}} & \textbf{DSC $\uparrow$} & \textbf{HD95 $\downarrow$} & \textbf{IoU-0.5 $\uparrow$} & \textbf{IoU-0.75 $\uparrow$} & \textbf{IoU-0.9 $\uparrow$} & \textbf{mAP $\uparrow$} \\
\midrule
Cellpose~\cite{stringer2021cellpose} 
    & \multicolumn{2}{l}{cyto3} 
    & 63.66$\pm$1.23 
    & 137.90$\pm$0.95 
    & 25.31$\pm$1.11 
    & 12.35$\pm$0.87 
    & 3.64$\pm$1.45 
    & 13.77$\pm$0.67 \\
    & \multicolumn{2}{l}{cyto2} 
    & 48.15$\pm$0.56 
    & 181.50$\pm$1.98 
    & 19.02$\pm$1.34 
    & 8.22$\pm$0.75 
    & 2.57$\pm$1.05 
    & 9.94$\pm$0.62 \\
\midrule
UNet++~\cite{zhou2019unet++_unet7} 
    & \multicolumn{2}{c}{-} 
    & 76.24$\pm$1.29 
    & 39.53$\pm$1.70 
    & 82.24$\pm$1.12 
    & 43.46$\pm$1.05 
    & 4.15$\pm$0.60 
    & 43.28$\pm$0.55 \\
\midrule
\rowcolor{gray!15} 
UNet++ + MorphoSeg 
    & \multicolumn{2}{c}{-} 
    & 76.49$\pm$1.27 
    & 39.29$\pm$1.68 
    & 82.68$\pm$1.15 
    & 44.09$\pm$1.00 
    & 4.47$\pm$0.57 
    & 43.75$\pm$0.54 \\
\midrule
SwinUnet~\cite{swinunet} 
    & \multicolumn{2}{c}{-} 
    & 72.78$\pm$1.48 
    & 42.77$\pm$1.95 
    & 77.29$\pm$1.10 
    & 31.36$\pm$0.95 
    & 3.27$\pm$0.65 
    & 37.31$\pm$0.53 \\
\midrule
\rowcolor{gray!15} 
SwinUnet + MorphoSeg 
    & \multicolumn{2}{c}{-} 
    & 75.23$\pm$1.42 
    & 39.47$\pm$1.83 
    & 81.13$\pm$1.07 
    & 39.51$\pm$1.02 
    & 3.91$\pm$0.70 
    & 41.52$\pm$0.55 \\
\midrule
ScaleFormer~\cite{Huang2022ScaleFormerRT} 
    & \multicolumn{2}{c}{-} 
    & 72.13$\pm$1.38 
    & 44.09$\pm$1.90 
    & 75.54$\pm$1.10 
    & 33.28$\pm$0.95 
    & 2.67$\pm$0.60 
    & 37.16$\pm$0.52 \\
\midrule
\rowcolor{gray!15} 
ScaleFormer + MorphoSeg 
    & \multicolumn{2}{c}{-} 
    & 75.11$\pm$1.35 
    & 41.11$\pm$1.80 
    & 80.29$\pm$1.08 
    & 39.03$\pm$0.92 
    & 2.95$\pm$0.55 
    & 40.76$\pm$0.51 \\
\midrule
TopoSeg~\cite{xu2025semi}           
    & \multicolumn{2}{l}{ViT-B16} 
    & 62.08$\pm$1.12 
    & 96.51$\pm$1.33 
    & 58.30$\pm$0.78 
    & 10.93$\pm$0.14 
    & 1.28$\pm$0.03 
    & 23.50$\pm$0.26 \\
    & \multicolumn{2}{l}{R50-B16} 
    & 66.14$\pm$1.55 
    & 97.77$\pm$1.23 
    & 64.21$\pm$1.67 
    & 15.12$\pm$0.78 
    & 1.83$\pm$1.11 
    & 27.05$\pm$0.72 \\
\midrule
\rowcolor{gray!15} 
TopoSeg + MorphoSeg 
    & \multicolumn{2}{l}{ViT-B16} 
    & 65.99$\pm$1.34 
    & 91.97$\pm$1.56 
    & 63.65$\pm$2.98 
    & 14.09$\pm$0.18 
    & 1.16$\pm$0.00 
    & 26.30$\pm$1.00 \\
\rowcolor{gray!15} 
    & \multicolumn{2}{l}{R50-B16} 
    & 67.97$\pm$1.11 
    & 91.40$\pm$1.45 
    & 68.56$\pm$1.23 
    & 17.24$\pm$0.56 
    & 2.19$\pm$1.67 
    & 29.33$\pm$0.72 \\
\midrule
TransUNet~\cite{chen2024transunet}
    & \multicolumn{2}{l}{ViT-B16} 
    & 72.97$\pm$1.50 
    & 32.94$\pm$1.20 
    & 78.33$\pm$1.11 
    & 34.52$\pm$0.87 
    & 3.51$\pm$0.89 
    & 38.79$\pm$0.56 \\
    & \multicolumn{2}{l}{R50-B16} 
    & 80.35$\pm$1.67 
    & 21.98$\pm$0.99 
    & 88.00$\pm$1.11 
    & 58.67$\pm$1.45 
    & 9.22$\pm$1.02 
    & 51.96$\pm$0.70 \\
\midrule
\rowcolor{gray!15} 
TransUNet + MorphoSeg 
    & \multicolumn{2}{l}{ViT-B16} 
    & 84.89$\pm$1.56 
    & \textbf{15.37$\pm$1.78} 
    & 92.50$\pm$1.23 
    & 74.02$\pm$1.36 
    & 18.08$\pm$0.12 
    & 61.53$\pm$0.61 \\
\rowcolor{gray!15}   
    & \multicolumn{2}{l}{R50-B16} 
    & \textbf{86.57$\pm$1.34} 
    & 15.75$\pm$1.11 
    & \textbf{93.02$\pm$2.45} 
    & \textbf{80.01$\pm$1.93} 
    & \textbf{21.87$\pm$0.67} 
    & \textbf{64.97$\pm$1.06} \\
\bottomrule
\end{tabular}

%% file: tab/abl_loss.tex
\begin{tabular}{ccccccccc}
\toprule
\textbf{Method}            & \textbf{Encoder} & \textbf{Loss}  & \textbf{DSC$\uparrow$}         & \textbf{HD95$\downarrow$}        & \textbf{IoU-0.5$\uparrow$}        & \textbf{IoU-0.75$\uparrow$}       & \textbf{IoU-0.9$\uparrow$}        & \textbf{mAP$\uparrow$}           \\
\midrule
\multirow{6}{*}{MorphoSeg} 
  & \multirow{3}{*}{ViT-B16} 
      & Balance  & 76.20$\pm$1.50  & 31.77$\pm$3.00  & 65.34$\pm$1.00   & 39.78$\pm$0.90   & 4.07$\pm$0.50   & 36.40$\pm$0.48  \\
  &                       & Norm     & 84.89$\pm$1.56  & \textbf{15.37$\pm$1.78}  & 92.50$\pm$1.23   & 74.02$\pm$1.36   & 18.08$\pm$0.12  & 61.53$\pm$0.61  \\
  &                       & Pareto   & 79.08$\pm$1.34  & 26.20$\pm$1.11  & 69.01$\pm$1.45   & 49.92$\pm$1.67   & 5.95$\pm$0.67   & 41.63$\pm$0.77  \\
\cmidrule(lr){2-9}
  & \multirow{3}{*}{R50-B16} 
      & Balance  & 86.33$\pm$1.11  & 16.33$\pm$0.99   & 93.02$\pm$1.45  & 79.68$\pm$1.42  & 20.11$\pm$1.67  & 64.27$\pm$0.88 \\
  &                       & Norm     & 84.40$\pm$1.16  & 26.37$\pm$1.11   & 91.64$\pm$1.22  & 72.50$\pm$1.67  & 11.30$\pm$1.34  & 58.48$\pm$0.82 \\
  &                       & Pareto   & \textbf{86.57$\pm$1.34} & 15.75$\pm$1.11 & \textbf{93.02$\pm$2.45} & \textbf{80.01$\pm$1.93} & \textbf{21.87$\pm$0.67} & \textbf{64.97$\pm$1.06} \\
\bottomrule
\end{tabular}

%% file: tab/abl_layers.tex
\begin{tabular}{cccccccc}
\toprule
\textbf{Encoder} & \textbf{Layers} & \textbf{Heads} & \textbf{Hybrid} & \textbf{Loss} & \textbf{DSC$\uparrow$} & \textbf{HD95$\downarrow$} & \textbf{mAP$\uparrow$} \\
\midrule
\multirow{3}{*}{ViT-B16} 
  & \multirow{3}{*}{12} 
  & \multirow{3}{*}{12} 
  & \multirow{3}{*}{$\times$} 
  & Balance & 76.20$\pm$1.50  & 31.77$\pm$3.00 & 42.31$\pm$0.69 \\
  &                      &                     &                     & Norm    & 84.89$\pm$1.56  & 15.37$\pm$1.78 & 61.53$\pm$0.61 \\
  &                      &                     &                     & Pareto  & 79.08$\pm$1.34  & 26.20$\pm$1.11 & 47.58$\pm$0.55 \\
\midrule
\multirow{3}{*}{ViT-L16} 
  & \multirow{3}{*}{24} 
  & \multirow{3}{*}{16} 
  & \multirow{3}{*}{$\times$} 
  & Balance & 77.35$\pm$1.45  & 30.12$\pm$2.85 & 43.27$\pm$0.74 \\
  &                      &                     &                     & Norm    & 85.42$\pm$1.59  & 14.88$\pm$1.77 & 62.34$\pm$0.68 \\
  &                      &                     &                     & Pareto  & 79.97$\pm$1.38  & 25.83$\pm$1.12 & 48.21$\pm$0.63 \\
\midrule
\multirow{3}{*}{R50-B16} 
  & \multirow{3}{*}{12} 
  & \multirow{3}{*}{12} 
  & \multirow{3}{*}{$\checkmark$} 
  & Balance & 86.33$\pm$1.11  & 16.33$\pm$0.99  & 64.27$\pm$0.77 \\
  &                      &                     &                     & Norm    & 84.40$\pm$1.16  & 26.37$\pm$1.11  & 58.48$\pm$0.64 \\
  &                      &                     &                     & Pareto  & 86.57$\pm$1.34  & 15.75$\pm$1.11  & 64.97$\pm$1.06 \\
\midrule
\multirow{3}{*}{R50-L32} 
  & \multirow{3}{*}{32} 
  & \multirow{3}{*}{16} 
  & \multirow{3}{*}{$\checkmark$} 
  & Balance & 88.50$\pm$1.26  & 14.50$\pm$1.72  & 67.32$\pm$0.71 \\
  &                      &                     &                     & Norm    & 87.55$\pm$1.15  & 17.82$\pm$1.19  & 62.75$\pm$0.68 \\
  &                      &                     &                     & Pareto  & \textbf{88.95$\pm$1.34}  & \textbf{14.25$\pm$1.21}  & \textbf{68.11$\pm$0.77} \\
\bottomrule
\end{tabular}

%% file: tab/dataset.tex
\begin{tabular}{lccccccc|cccccc}
\toprule
                           &  & \multicolumn{6}{c|}{\large{\textsc{BloodCell}\cite{Ljosa2012}}} & \multicolumn{6}{c}{\large{\textsc{LIVECell}\cite{livecell}}} \\
\cmidrule(lr){3-8}\cmidrule(lr){9-14}
\textbf{Method}  &   \textbf{Loss} & \textbf{DSC $\uparrow$} & \textbf{HD95 $\downarrow$} & \textbf{IoU-0.5 $\uparrow$} & \textbf{IoU-0.75 $\uparrow$} & \textbf{IoU-0.9 $\uparrow$} & \textbf{mAP $\uparrow$} & \textbf{DSC $\uparrow$} & \textbf{HD95 $\downarrow$} & \textbf{IoU-0.5 $\uparrow$} & \textbf{IoU-0.75 $\uparrow$} & \textbf{IoU-0.9 $\uparrow$} & \textbf{mAP $\uparrow$} \\
\midrule
UNet++\cite{zhou2019unet++_unet7}       & -                     & 96.11   & \textbf{2.21}   & \textbf{100.00}     & \textbf{100.00}   & 83.02    & 94.34    & 88.52   & 7.55   & \textbf{100.00}     & \textbf{71.93}   & \textbf{15.79}   & \textbf{62.57}  \\
\rowcolor{gray!15}    & Norm    & 95.98   & 2.26   & \textbf{100.00}     & \textbf{100.00}   & 79.25    & 93.08    & \textbf{88.88}   & \textbf{6.58}   & \textbf{100.00}     & 68.42   & 14.04   & 60.82  \\
\rowcolor{gray!15}   {+ MorphoSeg}                           & Balance & \textbf{96.13}   & 2.23   & \textbf{100.00}     & \textbf{100.00}   & \textbf{83.65}    & \textbf{94.55}    & 88.43   & 7.10   & \textbf{100.00}     & \textbf{71.93}   & \textbf{15.79}   & \textbf{62.57}  \\
\rowcolor{gray!15}                              & Pareto  & 96.07   & \textbf{2.21}   & \textbf{100.00}     & \textbf{100.00}   & 81.76    & 93.92    & 88.55   & 7.04   & \textbf{100.00}     & 68.42   & 12.28   & 60.23  \\
\midrule
TransUNet  \cite{chen2024transunet}      & -                     & 90.56   & 4.80   & \textbf{100.00}     & 98.74    & 0.00     & 66.25    & 73.83   & 18.08  & 77.19      & 19.30   & 0.00    & 32.16  \\
\rowcolor{gray!15}  & Norm    & \textbf{94.73}   & \textbf{2.66}   & \textbf{100.00}     & \textbf{99.37}    & \textbf{57.86}    & \textbf{85.74}    & \textbf{76.98}   & \textbf{16.27}  & \textbf{85.96}      & \textbf{24.56}   & \textbf{1.75}    & \textbf{37.43}  \\
\rowcolor{gray!15}       {+ MorphoSeg}                       & Balance & 90.57   & 4.75   & \textbf{100.00}     & 98.74    & 0.00     & 66.25    & 73.83   & 18.08  & 77.19      & 19.30   & 0.00    & 32.16  \\
\rowcolor{gray!15}                              & Pareto  & 92.16   & 4.02   & \textbf{100.00}     & 98.74    & 8.81     & 69.18    & 75.05   & 17.44  & 77.19      & 22.81   & \textbf{1.75}    & 33.92  \\
\midrule
SwinUNet   \cite{swinunet}  & -                     & 95.37   & 2.87   & \textbf{100.00}     & 99.37    & 69.81    & 89.73    & 82.95   & 9.37   & \textbf{100.00}     & 40.35   & 3.51    & 47.95  \\
\rowcolor{gray!15}  & Norm    & \textbf{95.95}   & \textbf{2.39}   & \textbf{100.00}     & \textbf{100.00}   & \textbf{81.13}    & \textbf{93.71}    & \textbf{85.33}   & 8.79   & \textbf{100.00}     & \textbf{50.88}   & \textbf{7.02}    & \textbf{52.63}  \\
\rowcolor{gray!15}  {+ MorphoSeg}       & Balance & 95.61   & 2.70   & \textbf{100.00}     & \textbf{100.00}   & 72.33    & 90.78    & 82.95   & 9.37   & \textbf{100.00}     & 40.35   & 3.51    & 47.95  \\
\rowcolor{gray!15}                              & Pareto  & 95.74   & 2.51   & \textbf{100.00}     & \textbf{100.00}   & 76.73    & 92.24    & 85.27   & \textbf{8.78}   & \textbf{100.00}     & \textbf{50.88}   & \textbf{7.02}    & \textbf{52.63}  \\
\midrule
ScaleFormer  \cite{Huang2022ScaleFormerRT}  & -                     & 95.91   & 2.17   & \textbf{100.00}     & \textbf{100.00}   & 78.62    & 92.87    & 88.48   & \textbf{6.52}   & \textbf{100.00}     & \textbf{71.93}   & 12.28   & 61.40  \\
\rowcolor{gray!15}  & Norm    & 96.04   & 2.16   & \textbf{100.00}     & \textbf{100.00}   & \textbf{83.02}    & \textbf{94.34}    & 83.24   & 10.97  & 96.49      & 43.86   & 3.51    & 47.95  \\
\rowcolor{gray!15} {+ MorphoSeg}   & Balance & \textbf{96.16}   & 2.20   & \textbf{100.00}     & \textbf{100.00}   & \textbf{83.02}    & \textbf{94.34}    & 87.89   & 8.41   & \textbf{100.00}     & 63.16   & 14.04   & 59.06  \\
\rowcolor{gray!15}                              & Pareto  & 96.04   & \textbf{2.15}   & \textbf{100.00}     & \textbf{100.00}   & 79.87    & 93.29    & \textbf{89.01}   & 7.95   & \textbf{100.00}     & 70.18   & \textbf{15.79}   & \textbf{61.99}  \\
\bottomrule
\end{tabular}

%% file: sec/6_conclusion.tex
\section{Conclusions and Future Works} \label{sec:con}
In this paper, we introduced a novel data repository containing differentiating NT2 cells, designed to facilitate the segmentation of complex cell morphologies. This dataset requires pixel-level segmentation, which presents new challenges for accurately segmenting cells with intricate shapes. To address these challenges, we proposed an uncertainty-aware segmentation method \model enhancing the base TranUNet by training on virtual outliers sampled near the feature representation space, thereby increasing the model's sensitivity to small pixel variations and morphological changes across different cell stages. This enhancement makes our \model more robust and effective for segmenting complex and irregularly shaped cells.


The experimental results demonstrate that the proposed \model outperforms existing benchmarks, including Cellpose and TransUNet. Our best-performing model, \model with a hybrid encoder, showed a notable improvement in performance metrics, achieving a 6.23\% increase in the Dice Similarity Coefficient (DSC). This statistic measures the similarity between the segmented cells and the ground truth. A reduction in HD95 (Hausdorff Distance at the 95th percentile) was also measured, indicating a decreased spatial discrepancy between the predicted and actual segmentation boundaries compared to the baseline configuration. By providing comprehensive visualizations and complete code, we aim to contribute to not only advance further research in this emerging domain but also to serve as a practical solution for cell segmentation and visualization in diverse medical AI applications.

\noindent\textbf{Limitations.} The relatively small size of the annotated dataset (36 images), is a limitation and the data may not be sufficient for all research applications. We anticipate that incorporating a larger number of images could further improve model performance. Additionally, while our method is effective for the NT2 cell dataset, it may not scale seamlessly to all types of cell datasets; inappropriately handling systematic outliers could potentially reduce its effectiveness.

\noindent\textbf{Future Work.} 
Future work will focus on a broader range of deep learning architectures to further enhance performance and generalisability. This also includes applying sampling procedures within the latent space to further simplify training and potentially enhance the results. Quantification of the method performance with respect to uncertainties is another important aspect to study further.




\section*{Acknowledgments}
We express our gratitude to the UK Engineering and Physical Sciences Research Council (EPSRC) for their support through the following projects: EP/V026747/1 (Trustworthy Autonomous Systems Node in Resilience) and EP/T013265/1 (NSF-EPSRC: ShiRAS. Towards Safe and Reliable Autonomy in Sensor Driven Systems). We also acknowledge the support of the National Science Foundation (NSF) under Grant No. ECCS 1903466 for the ShiRAS project. For the purpose of open access, the authors have applied a Creative Commons Attribution (CC BY) licence to any Author Accepted Manuscript (AAM) version arising from this work.

%% file: main.bbl
\begin{thebibliography}{10}
\expandafter\ifx\csname url\endcsname\relax
  \def\url#1{\texttt{#1}}\fi
\expandafter\ifx\csname urlprefix\endcsname\relax\def\urlprefix{URL }\fi
\expandafter\ifx\csname href\endcsname\relax
  \def\href#1#2{#2} \def\path#1{#1}\fi

\bibitem{chen2024transunet}
J.~Chen, J.~Mei, X.~Li, Y.~Lu, Q.~Yu, Q.~Wei, X.~Luo, Y.~Xie, E.~Adeli, Y.~Wang, et~al., {TransUNet: Rethinking the U-Net Architecture Design for Medical Image Segmentation Through the Lens of Transformers}, Medical Image Analysis (2024) 103280.

\bibitem{ronneberger2015u_unet0}
O.~Ronneberger, P.~Fischer, T.~Brox, {U-Net: Convolutional Networks for Biomedical Image Segmentation}, in: Proc. of the International Conference on Medical Image Computing and Computer-assisted Intervention, Springer, 2015, pp. 234--241.

\bibitem{zhang2023customized}
K.~Zhang, D.~Liu, Customized segment anything model for medical image segmentation, arXiv preprint arXiv:2304.13785 (2023).

\bibitem{zhao20213d}
T.~Zhao, K.~Cao, J.~Yao, I.~Nogues, L.~Lu, L.~Huang, J.~Xiao, Z.~Yin, L.~Zhang, {3D Graph Anatomy Geometry-Integrated Network For Pancreatic Mass Segmentation, Diagnosis, And Quantitative Patient Management}, in: Proceedings of the IEEE/CVF Conference on Computer Vision and Pattern Recognition, 2021, pp. 13743--13752.

\bibitem{saha2018her2net}
M.~Saha, C.~Chakraborty, {Her2Net: A Deep Framework for Semantic Segmentation and Classification of Cell Membranes and Nuclei in Breast Cancer Evaluation}, IEEE Transactions on Image Processing 27~(5) (2018) 2189--2200.

\bibitem{stringer2021cellpose}
C.~Stringer, T.~Wang, M.~Michaelos, M.~Pachitariu, {Cellpose: A Generalist Algorithm For Cellular Segmentation}, Nature methods 18~(1) (2021) 100--106.

\bibitem{616aacc35244ab9dcb31d91f}
V.~Petukhov, R.~J. Xu, R.~A. Soldatov, P.~Cadinu, K.~Khodosevich, J.~R. Moffitt, P.~V. Kharchenko, Cell segmentation in imaging-based spatial transcriptomics 40 (2021) 345--354.

\bibitem{livecell}
C.~Edlund, T.~R. Jackson, N.~Khalid, N.~Bevan, T.~Dale, A.~Dengel, S.~Ahmed, J.~Trygg, R.~Sjögren, {LIVECell---A large-scale dataset for label-free live cell segmentation}, Nature Methods 18~(9) (2021) 1038--1045.

\bibitem{israel2023foundation}
U.~Israel, M.~Marks, R.~Dilip, Q.~Li, M.~Schwartz, E.~Pradhan, E.~Pao, S.~Li, A.~Pearson-Goulart, P.~Perona, et~al., {CellSAM: A Foundation Model for Cell Segmentation}, bioRxiv.

\bibitem{10.1371/journal.pcbi.1005177}
D.~A. Van~Valen, T.~Kudo, K.~M. Lane, D.~N. Macklin, N.~T. Quach, M.~M. DeFelice, I.~Maayan, Y.~Tanouchi, E.~A. Ashley, M.~W. Covert, {Deep Learning Automates the Quantitative Analysis of Individual Cells in Live-Cell Imaging Experiments}, PLOS Computational Biology 12~(11) (2016) 1--24.

\bibitem{Moen803205}
E.~Moen, E.~Borba, G.~Miller, M.~Schwartz, D.~Bannon, N.~Koe, I.~Camplisson, D.~Kyme, C.~Pavelchek, T.~Price, T.~Kudo, E.~Pao, W.~Graf, D.~Van~Valen, {Accurate Cell Tracking and Lineage Construction in Live-Cell Imaging Experiments with Deep Learning}, bioRxiv (2019).

\bibitem{li2017h_unet1}
X.~Li, H.~Chen, X.~Qi, Q.~Dou, C.-W. Fu, P.~A. Heng, {H-DenseUNet: A Hybrid Densely Connected UNet for Liver and Liver Tumor Segmentation from CT Volumes}, IEEE Transactions on Medical Imaging 37~(12) (2018) 2663--2674.

\bibitem{5fe31675d4150a363c80bcfa}
T.~Zhao, Z.~Yin, {Weakly Supervised Cell Segmentation by Point Annotation}, IEEE Transactions on Medical Imaging 40~(10) (2021) 2736--2747.

\bibitem{63634e0790e50fcafd9fca3a}
M.~C. Robitaille, J.~M. Byers, J.~A. Christodoulides, M.~P. Raphael, {Self-supervised Machine Learning for Live Cell Imagery Segmentation}, Communications Biology 5 (2022) 1162--8.

\bibitem{data_repository:2024:sheffield}
H.~J. McCourty, T.~Zhang, A.~Nikolaev, B.~M. Sanchez-Tafoll, L.~S. Mihaylova, Dataset of {Ntera-2} cells for segmentation (2024).
\newblock \href {https://doi.org/https://doi.org/10.15131/shef.data.25604421} {\path{doi:https://doi.org/10.15131/shef.data.25604421}}.

\bibitem{pleasure1993}
S.~Pleasure, V.~Lee, {NTera 2 Cells: A Human Cell Line Which Displays Characteristics Expected of a Human Committed Neuronal Progenitor Cell}, Journal of Neuroscience Research 34 (1993) 585--602.

\bibitem{andrews1984}
P.~W. Andrews, {Retinoic acid induces neuronal differentiation of a cloned human embryonal carcinoma cell line in vitro}, Developmental Biology 103~(2) (1984) 285--293.

\bibitem{liu20173d_unet2}
S.~Liu, D.~Xu, S.~K. Zhou, O.~Pauly, S.~Grbic, T.~Mertelmeier, J.~Wicklein, A.~Jerebko, W.~Cai, D.~Comaniciu, {3D Anisotropic Hybrid Network: Transferring Convolutional Features from 2D Images to 3D Anisotropic Volumes}, in: International Conference on Medical Image Computing and Computer-Assisted Intervention, Springer, 2018, pp. 851--858.

\bibitem{milletari2016v_unet3}
F.~Milletari, N.~Navab, S.-A. Ahmadi, {V-Net}: Fully convolutional neural networks for volumetric medical image segmentation, in: 2016 Fourth International Conference on 3D Vision (3DV), IEEE, 2016, pp. 565--571.

\bibitem{yu2017volumetric_unet4}
L.~Yu, X.~Yang, H.~Chen, J.~Qin, P.-A. Heng, {Volumetric ConvNets with Mixed Residual Connections for Automated Prostate Segmentation from 3D MR Images}, in: Proceedings of the AAAI Conference on Artificial Intelligence, 2017.

\bibitem{zhou2019unet++_unet7}
Z.~Zhou, M.~M.~R. Siddiquee, N.~Tajbakhsh, J.~Liang, {Unet++: Redesigning Skip Connections to Exploit Multiscale Features in Image Segmentation}, IEEE Transactions on Medical Imaging 39~(6) (2019) 1856--1867.

\bibitem{isensee2021nnu}
F.~Isensee, P.~F. Jaeger, S.~A. Kohl, J.~Petersen, K.~H. Maier-Hein, {nnU-Net}: A self-configuring method for deep learning-based biomedical image segmentation, Nature methods 18~(2) (2021) 203--211.

\bibitem{vaswani2017attention}
A.~Vaswani, N.~Shazeer, N.~Parmar, J.~Uszkoreit, L.~Jones, A.~N. Gomez, L.~Kaiser, I.~Polosukhin, Attention is all you need, Advances in Neural Information Processing Systems 30 (2017).

\bibitem{dosovitskiy2020image}
A.~Dosovitskiy, L.~Beyer, A.~Kolesnikov, D.~Weissenborn, X.~Zhai, T.~Unterthiner, M.~Dehghani, M.~Minderer, G.~Heigold, S.~Gelly, et~al., An image is worth 16x16 words: Transformers for image recognition at scale, International Conference on Learning Representations (2021).

\bibitem{swinunet}
H.~Cao, Y.~Wang, J.~Chen, D.~Jiang, X.~Zhang, Q.~Tian, M.~Wang, {Swin-Unet: Unet-like Pure Transformer for Medical Image Segmentation}, in: Proceedings of the European Conference on Computer Vision Workshops(ECCVW), 2022.

\bibitem{Huang2022ScaleFormerRT}
H.~Huang, S.~Xie, L.~Lin, Y.~Iwamoto, X.~Han, Y.-W. Chen, R.~Tong, {ScaleFormer: Revisiting the Transformer-based Backbones from a Scale-wise Perspective for Medical Image Segmentation}, ArXiv abs/2207.14552 (2022).

\bibitem{fan2020pranet}
T.~Zhou, Y.~Zhou, K.~He, C.~Gong, J.~Yang, H.~Fu, D.~Shen, {Cross-level Feature Aggregation Network for Polyp Segmentation}, Pattern Recognition 140 (2023) 109555.

\bibitem{CellViT}
F.~Hörst, M.~Rempe, L.~Heine, C.~Seibold, J.~Keyl, G.~Baldini, S.~Ugurel, J.~Siveke, B.~Grünwald, J.~Egger, J.~Kleesiek, {CellViT: Vision Transformers for precise cell segmentation and classification}, Medical Image Analysis 94 (2024) 103143.

\bibitem{kirillov2023segment}
A.~Kirillov, E.~Mintun, N.~Ravi, H.~Mao, C.~Rolland, L.~Gustafson, T.~Xiao, S.~Whitehead, A.~C. Berg, W.-Y. Lo, et~al., Segment anything, arXiv preprint arXiv:2304.02643 (2023).

\bibitem{gupta2024topology}
S.~Gupta, Y.~Zhang, X.~Hu, P.~Prasanna, C.~Chen, {Topology-aware Uncertainty for Image Segmentation}, in: Advances in Neural Information Processing Systems, Vol.~36, 2024.

\bibitem{lin2024topout}
S.~Lin, T.~Sabyrbayev, Z.~Jin, G.~Li, H.~Cao, D.~Zou, {TopoUT: Segmentation Through Efficient Topological Regularization}, in: Proceedings of the IEEE International Symposium on Biomedical Imaging (ISBI), IEEE, 2024, pp. 1--5.

\bibitem{liu2024enhancing}
C.~Liu, B.~Ma, X.~Ban, Y.~Xie, H.~Wang, W.~Xue, K.~Xu, {Enhancing Boundary Segmentation for Topological Accuracy with Skeleton-Based Methods}, arXiv preprint arXiv:2404.18539 (2024).

\bibitem{xu2025semi}
M.~Xu, X.~Hu, S.~Gupta, S.~Abousamra, C.~Chen, {Semi-supervised Segmentation of Histopathology Images with Noise-aware Topological Consistency}, in: Proceedings of the European Conference on Computer Vision, Springer, Cham, 2025, pp. 271--289.

\bibitem{Aggarwal2017}
C.~C. Aggarwal, {An Introduction to Outlier Analysis}, Springer International Publishing, Cham, 2017, pp. 1--34.

\bibitem{wang2019progress}
H.~Wang, M.~J. Bah, M.~Hammad, {Progress in Outlier Detection Techniques: A Survey}, Ieee Access 7 (2019) 107964--108000.

\bibitem{boukerche2020outlier}
A.~Boukerche, L.~Zheng, O.~Alfandi, {Outlier Detection: Methods, Models, and Classification}, ACM Computing Surveys (CSUR) 53~(3) (2020) 1--37.

\bibitem{du2022towards}
X.~Du, Z.~Wang, M.~Cai, Y.~Li, Vos: Learning what you don’t know by virtual outlier synthesis, in: Proceedings of the International Conference on Learning Representations, 2022.

\bibitem{dong2022neural}
X.~Dong, J.~Guo, A.~Li, W.-T. Ting, C.~Liu, H.~Kung, Neural mean discrepancy for efficient out-of-distribution detection, in: Proceedings of the IEEE/CVF Conference on Computer Vision and Pattern Recognition, 2022, pp. 19217--19227.

\bibitem{BOWLES2017643}
C.~Bowles, C.~Qin, R.~Guerrero, R.~Gunn, A.~Hammers, D.~A. Dickie, M.~Vald{\'e}s~Hern{\'a}ndez, J.~Wardlaw, D.~Rueckert, Brain lesion segmentation through image synthesis and outlier detection, NeuroImage: Clinical 16 (2017) 643--658.

\bibitem{thambawita2022singan}
V.~Thambawita, P.~Salehi, S.~A. Sheshkal, S.~A. Hicks, H.~L. Hammer, S.~Parasa, T.~d. Lange, P.~Halvorsen, M.~A. Riegler, {SinGAN-Seg: Synthetic training data generation for medical image segmentation}, PloS one 17~(5) (2022) e0267976.

\bibitem{Nie2020}
D.~Nie, D.~Shen, {Adversarial Confidence Learning for Medical Image Segmentation and Synthesis}, International Journal of Computer Vision 128~(10) (2020) 2494--2513.

\bibitem{5a260c8117c44a4ba8a30b08}
T.-Y. Lin, P.~Goyal, R.~Girshick, K.~He, P.~Dollár, {Focal Loss for Dense Object Detection}, in: 2017 IEEE International Conference on Computer Vision (ICCV), 2017, pp. 2999--3007.

\bibitem{tian2024fairseg}
Y.~Tian, M.~Shi, Y.~Luo, A.~Kouhana, T.~Elze, M.~Wang, {FairSeg: A Large-Scale Medical Image Segmentation Dataset for Fairness Learning Using Segment Anything Model with Fair Error-Bound Scaling}, in: International Conference on Learning Representations (ICLR), 2024.

\bibitem{shamir2019continuous}
R.~R. Shamir, Y.~Duchin, J.~Kim, G.~Sapiro, N.~Harel, {Continuous Dice Coefficient: A Method for Evaluating Probabilistic Segmentations}, arXiv preprint arXiv:1906.11031 (2019).

\bibitem{huttenlocher1993comparing}
D.~P. Huttenlocher, G.~A. Klanderman, W.~J. Rucklidge, {Comparing Images Using the Hausdorff Distance}, IEEE Transactions on Pattern Analysis and Machine Intelligence 15~(9) (1993) 850--863.

\bibitem{rezatofighi2019generalized}
H.~Rezatofighi, N.~Tsoi, J.~Gwak, A.~Sadeghian, I.~Reid, S.~Savarese, {Generalized Intersection Over Union: A Metric and a Loss for Bounding Box Regression}, in: Proceedings of the IEEE/CVF Conference on Computer Vision and Pattern Recognition, 2019, pp. 658--666.

\bibitem{henderson2017end}
P.~Henderson, V.~Ferrari, {End-to-End Training of Object Class Detectors for Mean Average Precision}, in: Computer Vision--ACCV 2016: 13th Asian Conference on Computer Vision, Taipei, Taiwan, November 20-24, 2016, Revised Selected Papers, Part V 13, Springer, 2017, pp. 198--213.

\bibitem{NeurIPS2018_Sener_Koltun}
O.~Sener, V.~Koltun, Multi-task learning as multi-objective optimization, in: S.~Bengio, H.~Wallach, H.~Larochelle, K.~Grauman, N.~Cesa-Bianchi, R.~Garnett (Eds.), Advances in Neural Information Processing Systems 31, Curran Associates, Inc., 2018, pp. 525--536.

\bibitem{Ljosa2012}
V.~Ljosa, K.~L. Sokolnicki, A.~E. Carpenter, {Annotated High-throughput Microscopy Image Sets for Validation}, Nature Methods 9~(7) (2012) 637--637.

\end{thebibliography}
